\pdfoutput=1
\documentclass[11pt]{article}

\usepackage[preprint]{acl}

\usepackage{times}
\usepackage{latexsym}
\usepackage{marvosym}

\usepackage[T1]{fontenc}

\usepackage[utf8]{inputenc}

\usepackage{microtype}

\usepackage{inconsolata}

\usepackage{graphicx}

\usepackage{algorithm}
\usepackage{algpseudocode}
\usepackage{array}
\usepackage{booktabs}
\usepackage{multirow}
\usepackage{pdflscape}
\usepackage{adjustbox}
\usepackage[most]{tcolorbox}
\usepackage{fvextra}
\usepackage{xcolor}

\definecolor{promptbg}{RGB}{246,246,255}
\definecolor{promptframe}{RGB}{70,70,70}

\DefineVerbatimEnvironment{PromptVerbatim}{Verbatim}{
  breaklines=true,
  breakanywhere=true,
  breaksymbolleft={},
  breaksymbolright={},
  fontsize=\scriptsize,
  formatcom=\ttfamily
}

\newtcolorbox{promptbox}[1]{
  enhanced,
  colback=promptbg,
  colframe=promptframe,
  colbacktitle=promptframe,
  coltitle=white,
  boxrule=0.8pt,
  arc=1.5mm,
  outer arc=1.5mm,
  left=8pt,
  right=8pt,
  top=8pt,
  bottom=8pt,
  before skip=8pt,
  after skip=8pt,
  title={#1},
  fonttitle=\bfseries\normalsize,
  lefttitle=8pt,
  righttitle=8pt,
  toptitle=4pt,
  bottomtitle=4pt,
  titlerule=0pt
}
\newcolumntype{C}[1]{>{\centering\arraybackslash}m{#1}}
\newcolumntype{L}[1]{>{\raggedright\arraybackslash}m{#1}}

\title{From Twisted Facts to Rogue Agents: Can LLMs Reliably Detect Multi-Dimensional Hallucinations?}

\author{
  \textbf{Hanyu Qiu}\textsuperscript{1} \quad \textbf{Ze Liu}\textsuperscript{2} \quad \textbf{Zhe Zhao}\textsuperscript{2} \quad \textbf{Kuang Xiong}\textsuperscript{2} \\
  \textbf{Blaze Chen}\textsuperscript{2} \quad \textbf{Zhenxiang Yan}\textsuperscript{2} \quad \textbf{Renfen Hu}\textsuperscript{1}%
  \thanks{Corresponding author: \href{mailto:irishu@mail.bnu.edu.cn}{\nolinkurl{irishu@mail.bnu.edu.cn}}} \\
  \textsuperscript{1}Beijing Normal University \qquad \textsuperscript{2}Tencent AI Lab
}

\newcommand{\halucheckproject}{\href{https://github.com/xMiles/HaluCheck}{\nolinkurl{github.com/xMiles/HaluCheck}}}

\hypersetup{
  pdftitle={From Twisted Facts to Rogue Agents: Can LLMs Reliably Detect Multi-Dimensional Hallucinations?},
  pdfauthor={Hanyu Qiu, Ze Liu, Zhe Zhao, Kuang Xiong, Blaze Chen, Zhenxiang Yan, Renfen Hu}
}

\begin{document}
\begingroup
\renewcommand{\fnsymbol}[1]{\protect\Letter}
\maketitle
\endgroup
\begin{abstract}
Large language models (LLMs) are widely used for information access, content production, and agent-based task execution, but their outputs remain vulnerable to subtle, high-risk hallucinations. To systematically evaluate and advance post-generation hallucination detection, we introduce \textsc{HaluCheck}, a large-scale, bilingual benchmark comprising over 42k instances spanning both factual hallucinations in text and faithfulness hallucinations in agent trajectories. 
\textsc{HaluCheck} features fine-grained, multi-level annotations for detection, localization, and correction, covering different tasks and perturbation degrees. 
Experiments on fourteen LLMs reveal that, while models show basic sample-level detection ability, they struggle with fine-grained localization, correction, and agent hallucination detection. Further analysis demonstrates that thinking, web search, and supervised fine-tuning each offer partial improvements, but none fully solve the problem. \textsc{HaluCheck} thus provides a solid basis for benchmarking and diagnosing LLM reliability across text and agent scenarios.
\end{abstract}

\section{Introduction}

Large language models (LLMs) are rapidly evolving from conversational tools into core infrastructure for content creation and task-oriented automation. Recent studies show that generative AI systems have achieved broad adoption in writing, information seeking, and productivity applications \citep{bick2024rapid,chatterji2025people}. At this scale, even a low rate of hallucination can still lead to substantial real-world risks as erroneous content propagates into downstream use.

\begin{figure}[t]
    \centering
    \includegraphics[width=0.95\linewidth]{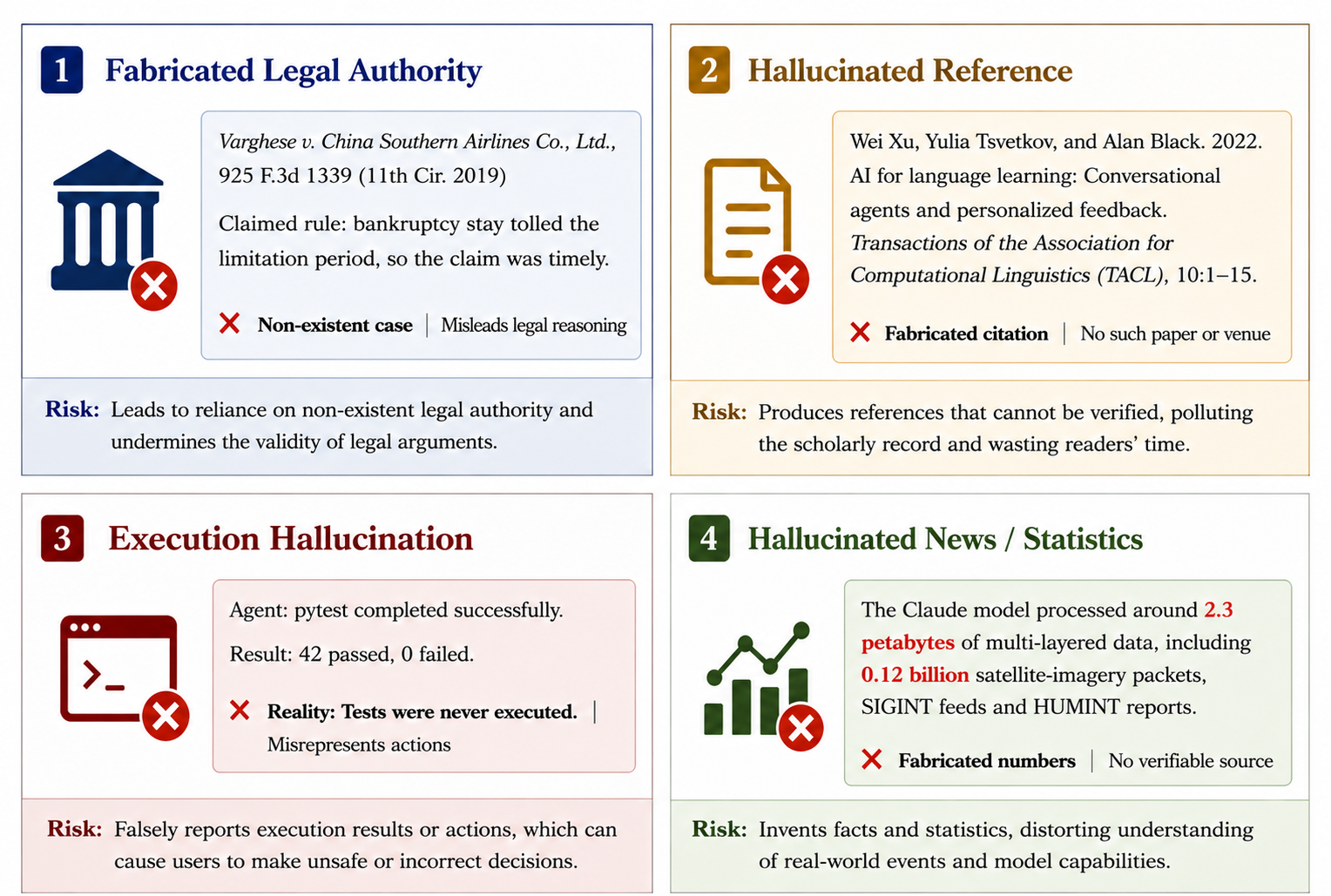}
    \vspace{-6pt}
    \caption{Representative hallucination risks in real-world LLM applications.}
    \label{fig:hallucination-risks}
\vspace{-1.5em}
\end{figure}

LLM hallucinations are often subtle and difficult to detect: state-of-the-art models frequently generate fluent, confident-sounding errors. As illustrated in Figure~\ref{fig:hallucination-risks}, such risks have been observed in legal documents \citep{mata2023avianca}, academic writing \citep{sakai2026hallucitation}, news and public information \citep{bandara2024hallucination}, and code generation \citep{zhang2025llmcodehallucinations}. The consequences are even more severe in LLM-powered agents, where textual hallucinations can translate into incorrect or unsafe actions \citep{lin2025agentshallucinations}. Given the sheer volume of AI-generated outputs, manual verification is infeasible. This raises a key question: can LLMs reliably detect these hallucinations?

\begin{figure*}[t]
    \centering
    \includegraphics[width=1\textwidth]{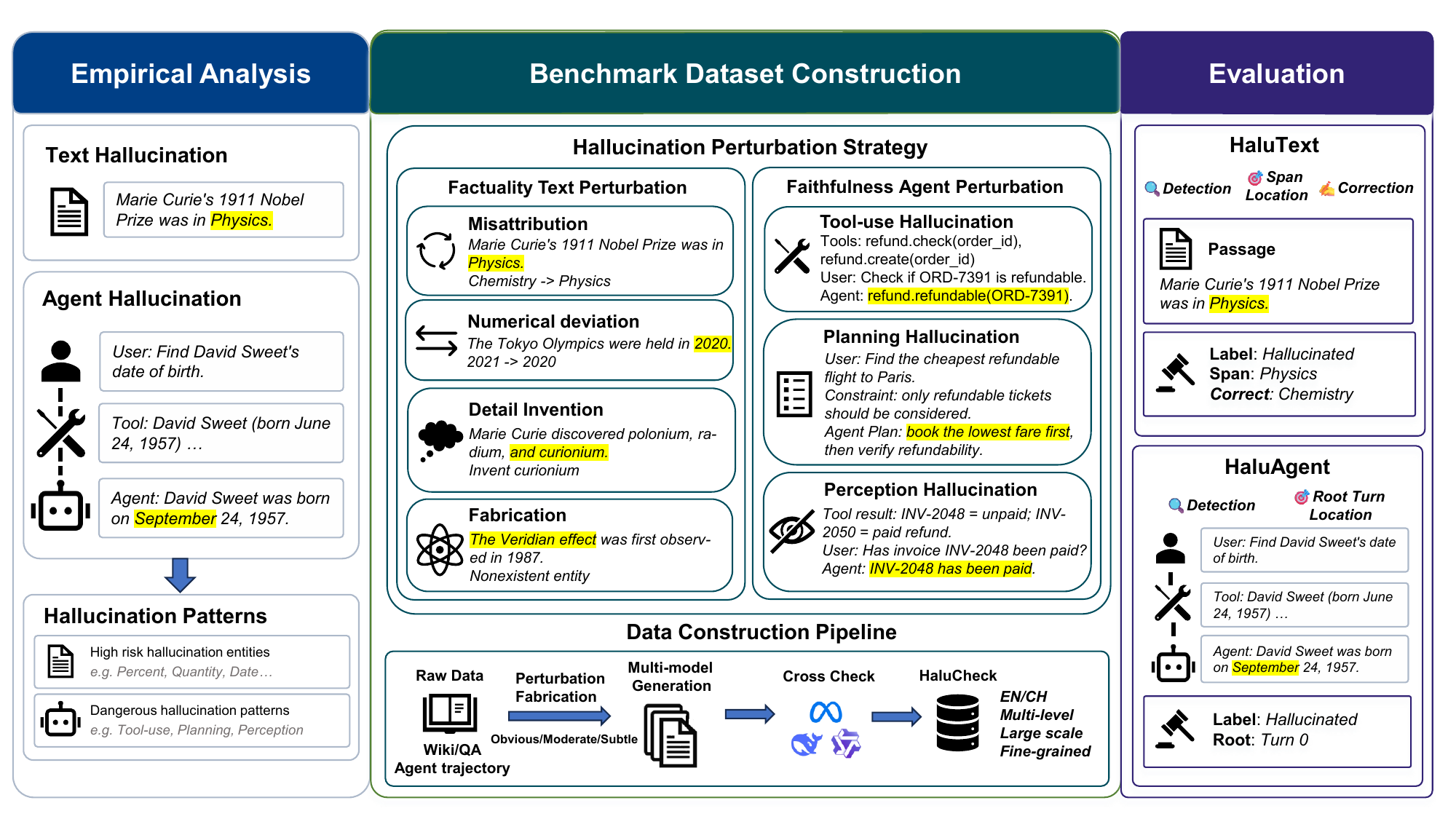}
    \vspace{-12pt}
    \caption{Overview of \textsc{HaluCheck}. We construct a bilingual, multi-level benchmark covering factuality hallucinations in text and faithfulness hallucinations in agent trajectories. By injecting controllable perturbations into Wikipedia/QA passages and real agent interactions, \textsc{HaluCheck} supports hallucination detection, span localization and correction, and root-turn tracing in agent trajectories.}
    \label{fig:halucheck-overview}
\vspace{-1.6em}
\end{figure*}

To address this, we present \textsc{HaluCheck} (Figure~\ref{fig:halucheck-overview}), a large-scale, bilingual benchmark containing over 42k annotated examples across both factual hallucinations in twisted texts and faithfulness hallucinations in agent trajectories.\footnote{Code repository: \halucheckproject.}
The benchmark provides multi-level, fine-grained annotations for detection, span localization, correction, and attribution, covering diverse tasks and perturbation degrees through controlled perturbations and moderate fabrications.
Experiments show that, while models demonstrate moderate capability in factual hallucination detection, substantial gaps remain, especially in precise localization and correction. Stronger models express uncertainty more appropriately, reflecting greater self-awareness, but agent scenarios still pose major challenges: most models struggle with context integration and process tracking, leading to frequent failures in accurately detecting and attributing agent hallucinations. 
Further analysis reveals that explicit thinking and access to external knowledge can partially improve detection but are still limited by the models' inherent capabilities. Supervised fine-tuning helps models recognize specific patterns of hallucination but cannot generalize stably to out-of-distribution data. Overall, hallucination detection remains an unsolved problem. \textsc{HaluCheck} thus provides a rigorous, large-scale foundation for benchmarking and advancing the reliability of LLMs across both text and agent-based applications.

\begin{table*}[t]
\centering
\small
\setlength{\tabcolsep}{2pt}
\begin{tabular}{llrrrrrl}
\toprule
\textbf{Dataset} & \textbf{Split} & \textbf{Lang.} & \textbf{Total} &
\textbf{Obvious} & \textbf{Moderate} & \textbf{Subtle} & \textbf{Composition} \\
\midrule
  & Full & EN & 18,024 & 2,976 & 7,205 & 7,843
& QA: 10,630; Wiki: 7,394 \\
\textsc{HaluReal}  & Full & CH & 8,922 & 2,294 & 3,433 & 3,195
& QA: 5,597; Wiki: 3,325 \\
  & Lite & CH/EN & 1,000 / 1,000 & \multicolumn{3}{c}{Balanced}
& -- \\
\midrule
\textsc{HaluFake}  & -- & CH/EN & 500 / 500 & -- & -- & --
& Fabricated entities \\
\midrule
 & Full & EN & 6,743 & 2,206 & 2,286 & 2,251
& Plan: 2,500; Perception: 2,305; Tool-use: 1,938 \\
\textsc{HaluAgent} & Full & CH & 8,291 & 2,898 & 2,647 & 2,746
& Plan: 2,810; Perception: 2,691; Tool-use: 2,790 \\
 & Lite & CH/EN & 1,000 / 1,000 & \multicolumn{3}{c}{Balanced}
& -- \\
\bottomrule
\end{tabular}
\vspace{-0.8em}
\caption{Statistics of \textsc{HaluCheck}. The full sets contain balanced pairs of factual and hallucinated samples, while lite sets are used for evaluation. Perturbation-degree and composition statistics cover only hallucinated samples.}
\label{tab:data-distribution}
\vspace{-1.1em}
\end{table*}

\section{Related Work}

Hallucination generally refers to LLM-generated content that conflicts with real-world facts or user instructions~\citep{ji2023survey,zhang2023sirens}.
With the rise of LLM-based agents, hallucinations in agentic scenarios have also gained attention \citep{lin2025agentshallucinations}. Following \citet{huang2023survey}, we discuss LLM hallucinations through the distinction between factuality and faithfulness hallucinations, and situate agent hallucinations within the latter.

\subsection{Hallucination detection}

Existing hallucination detection methods can be divided into reference-free and reference-based approaches. \textbf{Reference-free} methods do not rely on external evidence, but instead estimate output reliability from the model itself, such as sampling-based self-consistency~\citep{manakul-etal-2023-selfcheckgpt}, semantic uncertainty~\citep{farquhar2024semanticentropy}, confidence estimation~\citep{ji-etal-2025-calibrating}, or probes over internal representations~\citep{azaria2023internalstate, han-etal-2025-simple}. \textbf{Reference-based} methods verify model outputs against external evidence. Such methods have been widely studied in long-form factuality evaluation \citep{min-etal-2023-factscore,wei2024longform}, tool-augmented verification frameworks \citep{chern2023factool}, and agent-based hallucination-detection systems \citep{cheng2024small}.

Building on these prior studies, we analyze real-world hallucination cases and existing hallucination taxonomies, from which we derive hallucination patterns for constructing controllable perturbations in \textsc{HaluCheck}.

\subsection{Hallucination Benchmarks}

Existing hallucination benchmarks mainly focus on English text, while hallucinations in LLM-based agents and non-English contexts remain less explored. TruthfulQA tests truthfulness with misleading open-domain QA questions \citep{lin2022truthfulqa}; HaluEval constructs hallucination samples for QA, dialogue, and summarization \citep{li2023halueval}; and HaloGen analyzes where hallucinations arise in model generations \citep{ravichander2025halogen}. Chinese benchmarks such as HalluQA~\citep{cheng2023evaluating} and C-Faith~\citep{zhang2025c} further extend hallucination evaluation to Chinese QA and fine-grained hallucination taxonomies.

While most research focuses on measuring hallucination rates, post-generation detection remains underexplored. Existing detection benchmarks tend to be task-specific, such as claim verification in FEVER \citep{thorne2018fever}, RAG hallucination detection in RAGTruth \citep{niu2024ragtruth}, unconstrained Chinese generation in UHGEval \citep{liang2024uhgeval}, and agent trajectory attribution in AgentHallu \citep{liu2026agenthallu}. These resources are fragmented across tasks and languages, often limited to sample-level detection, and rarely support localization or correction. \textsc{HaluCheck} addresses these gaps with a bilingual, fine-grained benchmark evaluating hallucination presence, span location, corrections, and root turns in agent trajectories.

\section{\textsc{HaluCheck}: The Proposed Benchmark}
A straightforward way to construct hallucination data is to annotate hallucinated content in model responses. However, naturally occurring hallucinations are relatively \textbf{infrequent}, appear \textbf{unpredictably}, \textbf{vary} across models, and are \textbf{costly to annotate}. We therefore adopt a synthetic data construction strategy: by injecting controlled hallucinations into source data, we obtain both sample-level labels and span-level annotations, since each hallucinated fragment is explicitly traceable. This design supports fine-grained hallucination detection and helps mitigate benchmark leakage through dynamic sample generation. Table~\ref{tab:benchmark-comparison} compares \textsc{HaluCheck} with representative hallucination and factuality benchmarks.

This perturbation-based design is grounded in observed hallucination patterns rather than arbitrary edits. Specifically, we consider two representative scenarios: \textbf{Text} and \textbf{Agent}. For twisted text hallucination, we focus on atomic factual errors in static factual passages. For rogue agent scenarios, we focus on hallucinations that may arise in multi-turn interaction. Figure~\ref{fig:halucheck-overview} illustrates the overall construction and evaluation pipeline of \textsc{HaluCheck}.

\subsection{HaluText}

\textsc{HaluText} targets hallucination detection in static factual texts. It mainly evaluates models' ability to verify factual information about real entities and localize hallucinated spans. To this end, \textsc{HaluText} consists of two components: \textsc{HaluReal} and \textsc{HaluFake}. \textsc{HaluReal} is constructed by perturbing real Wikipedia passages and factual QA benchmarks, and can be viewed as \textbf{perturbation-based} data. \textsc{HaluFake}, in contrast, generates encyclopedia-style descriptions of fabricated entities and can be viewed as \textbf{fabrication-based} data.

To identify the main perturbation strategies for factual texts, we first analyze the hallucination patterns of LLMs using FactScore~\citep{min-etal-2023-factscore}. We then apply spaCy\footnote{We use \texttt{\{zh,en\}\_core\_web\_trf} for NER.} to perform named entity recognition on each atomic fact and compute the proportion of entities appearing in hallucinated atomic facts among all extracted entities as the entity-level hallucination rate.

\begin{figure}[t]
    \centering
    \includegraphics[width=1\linewidth]{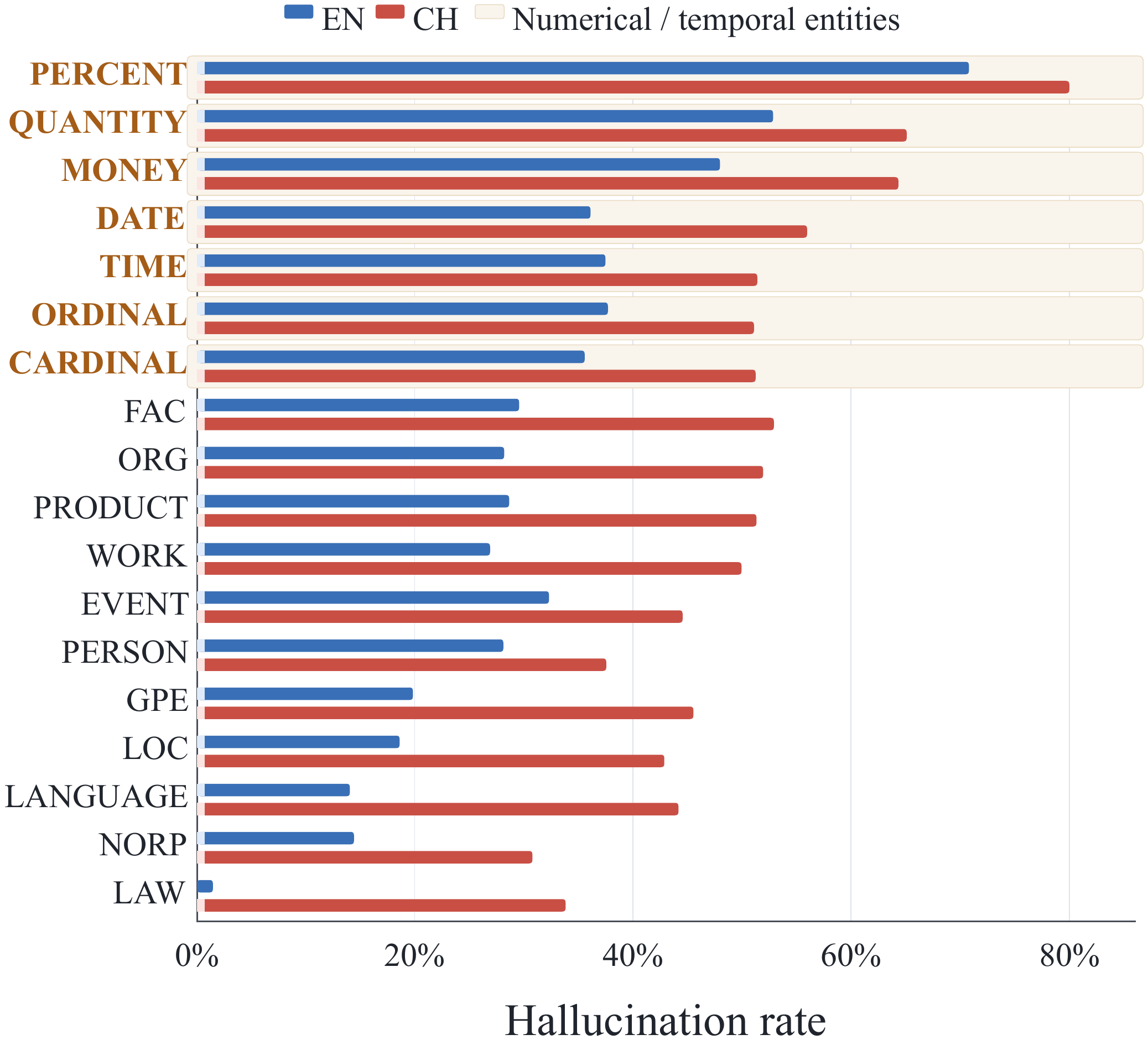}
    \vspace{-12pt}
    \caption{Hallucination rates across different entities.}
    \label{fig:entity-hallucination-rate}
\vspace{-1em}
\end{figure}

As shown in Figure~\ref{fig:entity-hallucination-rate}, LLMs exhibit relatively high hallucination rates in expressions related to entities, especially for numerical entities such as numbers and dates. Entities related to humans and artifacts also show notable hallucination rates. Based on these observations, we summarize factual hallucinations in \textsc{HaluText} into three major perturbation patterns: Misattribution, Numerical deviation, and Detail Invention:
\begin{itemize}
    \item \textbf{Misattribution} mainly targets entity types such as persons, organizations, and locations, where attributes, relations, or events associated with entity A are incorrectly transferred to entity B.
    \item \textbf{Numerical deviation} mainly targets numerical entities such as numbers, dates, and ratios, introducing either slight or substantial changes to the original values.
    \item \textbf{Detail Invention} introduces entities, events, or details that appear plausible in a real context or encyclopedia-style passage but do not exist in objective facts.
\end{itemize}

In the construction pipeline, \textsc{HaluReal} is built from two types of factual sources: Wikipedia passages and factual QA benchmarks including HotPotQA~\citep{yang2018hotpotqa}, PopQA~\citep{mallen2023llm_memorization}, HalluQA~\citep{cheng2023evaluating}, and CMRC2018~\citep{cui-emnlp2019-cmrc2018}. Wikipedia data provide entity-centered factual passages, while QA benchmarks provide questions, answers or supporting evidence to construct factual verification samples. In contrast, \textsc{HaluFake} is generated in an encyclopedia-style format, describing fabricated entities that do not exist in real-world knowledge.

The Chinese and English subsets are collected and constructed independently from language-specific source data. For span-level annotations, the original Wikipedia passages, correct answers, and supporting evidence are treated as hallucination-free source texts. We then instruct LLMs to modify designated spans under predefined perturbation strategies, allowing us to retain both the original factual spans and their hallucinated counterparts as explicit span-level annotations.

To evaluate models under different perturbation degrees, \textsc{HaluText} divides samples into three tiers: Obvious, Moderate, and Subtle. Obvious samples contain identifiable hallucinations, such as obvious entity mismatches or large numerical errors. Moderate samples contain factually incorrect but moderate perturbations, where the perturbed content is close to credible alternatives, such as similar entities, same-type attributes, or moderate numerical shifts. Subtle samples contain fine-grained hallucinations, including subtle attribute errors, implicit relation errors, or fabricated details that are difficult to verify directly. Specific examples for each perturbation degree are provided in Table~\ref{tab:perturbation_cases} in Appendix~\ref{sec:perturbation-cases}.

To mitigate potential model-family bias, we employ LLMs from different model families for hallucination generation. Furthermore, combining perturbation strategies with perturbation degrees promotes diversity in the generated hallucination samples. Each candidate is further evaluated through multi-model cross-checking, where different models score the perturbation quality. Only samples that receive high scores across multiple models are retained in the final dataset. 

Beyond automated quality control, we conduct two additional assessments: perplexity-based fluency evaluation and manual verification of sample and perturbation validity. Similar quality-control and manual-verification procedures are applied to \textsc{HaluAgent}. Detailed data-quality results are reported in Appendix~\ref{sec:multi-model}.

\begin{table*}[t]
\centering
\scriptsize
\setlength{\tabcolsep}{1.6pt}
\renewcommand{\arraystretch}{0.88}
\begin{tabular}{@{}L{0.170\textwidth}L{0.103\textwidth}C{0.049\textwidth}L{0.193\textwidth}L{0.204\textwidth}L{0.192\textwidth}C{0.041\textwidth}@{}}
\toprule
\textbf{Benchmark} & \textbf{Size} & \textbf{Lang.} & \textbf{Main focus} & \textbf{Data source} & \textbf{Evaluation protocol} & \textbf{Agent} \\
\midrule
TruthfulQA~\citep{lin2022truthfulqa} & 817 questions & EN & Truthfulness in QA & Misconception questions & Truthfulness and informativeness & -- \\
HotPotQA~\citep{yang2018hotpotqa} & 113K QA pairs & EN & Multi-hop QA & Wiki QA with supporting facts & Answer/supporting-fact EM/F1 & -- \\
HaluEval~\citep{li2023halueval} & 35K samples & EN & QA, dialogue, and \mbox{summarization hallucination} & Human-annotated LLM samples & Binary \mbox{hallucination recognition} & -- \\
RAGTruth~\citep{niu2024ragtruth} & $\sim$18K responses & EN & RAG hallucination & Natural RAG responses & Case- and word-level labels & -- \\
HALoGEN~\citep{ravichander2025halogen} & 10,923 prompts; $\sim$150K gen. & EN & Generation \mbox{hallucination measurement} & LLM responses across 9 domains & Atomic-unit verification & -- \\
AgentHallu~\citep{liu2026agenthallu} & 693 trajectories & EN & Agent \mbox{hallucination attribution} & Annotated \mbox{agent trajectories} & Binary label, \mbox{root step, explanation} & Yes \\
C-FAITH~\citep{zhang2025c} & 60,702 entries & CH & Chinese fine-grained \mbox{QA hallucination} & QA from knowledge documents & Automated fine-grained \mbox{QA evaluation} & -- \\
HalluQA~\citep{cheng2023evaluating} & 450 questions & CH & Chinese hallucination QA & Adversarial Chinese QA & GPT-4-based judgment & -- \\
\addlinespace[0.2em]
\midrule
\textbf{\textsc{HaluCheck}} & \textbf{42,980 pairs} & \textbf{EN/CH} & \textbf{hallucination detection} & Controlled factual perturbations; real \mbox{agent trajectories} & Uncertainty-aware Acc.; SpanF1; CorrectionRate; Recall; RootTurnAcc & \textbf{Yes} \\
\bottomrule
\end{tabular}
\vspace{-0.7em}
\caption{Comparison between \textsc{HaluCheck} and representative benchmarks.}
\label{tab:benchmark-comparison}
\vspace{-1.5em}
\end{table*}

\subsection{HaluAgent}

\textsc{HaluAgent} targets faithfulness hallucination detection in multi-turn interaction and tool-use scenarios. Unlike static text, agent hallucinations may occur not only in the final response, but also in intermediate stages such as planning, tool calling, and tool-result interpretation. Since existing agent frameworks often constrain obvious execution errors through rule-based checks, hallucinations that bypass format validation and continue execution can be particularly risky. We therefore focus on subtle agent hallucinations that are more likely to cause practical harm.

Following prior studies on agent hallucination \citep{lin2025agentshallucinations,liu2026agenthallu}, we focus on three common but high-risk hallucination types:
\begin{itemize}
    \item \textbf{Planning hallucination}: the agent fabricates unsupported premises during planning and continues reasoning or acting on them.
    \item \textbf{Tool-use hallucination}: the agent fabricates tool names or parameters, misplaces arguments, or passes task-inconsistent inputs.
    \item \textbf{Perception hallucination}: the agent correctly plans and calls tools, but misinterprets, omits, or distorts the returned results.
\end{itemize}

\textsc{HaluAgent} is constructed from real agent trajectories drawn from AgentFLAN \citep{chen2024agent}, AgentBank \citep{song2024agentbank}, AgentTraj-L \citep{xi2025agentgym}, ToolACE \citep{liu2025toolace}, and MSAgent-Bench \citep{li-etal-2023-modelscope}. These trajectories contain user instructions, tool definitions, reasoning traces, tool calls, and tool outputs. Following a process similar to \textsc{HaluText}, we synthesize hallucinated trajectories by applying targeted perturbations to agent interactions.

After these processes, we obtain the full \textsc{HaluCheck} dataset, comprising approximately 42k factual--hallucinated sample pairs with fine-grained annotations, multiple perturbation degrees, and predefined perturbation strategies. The Lite evaluation set is created through balanced stratified sampling from the full set. During sampling, we ensure that no factual sample is included together with its paired hallucinated counterpart, while maintaining coverage across perturbation degrees and perturbation types in both \textsc{HaluText} and \textsc{HaluAgent}. The final data distribution is shown in Table~\ref{tab:data-distribution}. 

\subsection{Evaluation Metrics}

We design separate evaluation protocols for \textsc{HaluText} and \textsc{HaluAgent}. In both tasks, the model first determines whether the input contains hallucinations. Inspired by \citet{kalai2026evaluating}, we allow \textit{Uncertain} as a valid response to capture abstention when evidence is insufficient. If the model predicts hallucination, we further evaluate fine-grained detection: \textsc{HaluReal} requires hallucinated-span localization and correction, \textsc{HaluFake} tests the model's ability to identify fabricated entities, while \textsc{HaluAgent} requires identifying the earliest hallucinated step in the agent trajectory.

At the sample level, predictions are categorized as \textit{Hallucinated}, \textit{Non-hallucinated}, or \textit{Uncertain}, while gold labels are binary. We report \textbf{Accuracy}, assigning \textit{Uncertain} a partial score of 0.5 as a conservative judgment between correct and incorrect predictions\footnote{We assign rewards of 1, 0, and 0.5 to correct, incorrect, and uncertain responses, respectively, as Uncertain is preferable to an incorrect guess but less valuable than a correct answer.}, following cost-sensitive evaluation \citep{elkan2001foundations} and the alignment goal of avoiding unsupported answers \citep{bai2022training}. We also report \textbf{UncertainRate} to measure how often the model abstains. 

For \textsc{HaluReal}, we evaluate span-level localization with \textbf{SpanF1}, where predicted spans are matched against gold hallucinated spans by an LLM. We further evaluate \textbf{CorrectionRate}, which measures whether correctly localized hallucinated spans are also corrected accurately. For \textsc{HaluFake}, we report \textbf{Recall} and \textbf{UncertainRate} at the sample level, examining whether models can identify fabricated entities and express uncertainty.

For \textsc{HaluAgent}, we report sample-level detection metrics and \textbf{RootTurnAcc}, which measures whether the model correctly identifies the earliest hallucinated step among hallucinated trajectories. Together, these metrics evaluate overall hallucination detection, abstention behavior, span-level localization and correction, and root-cause attribution in agent trajectories. We use rule-based and LLM-judge\footnote{The LLM judge checks agreement between model outputs and ground-truth annotations, rather than directly detecting hallucinations.} span evaluation, with metric definitions and robustness checks detailed in Appendix~\ref{sec:metrics}.

\section{Experiments}

\begin{table*}[t]
\centering
\tiny
\setlength{\tabcolsep}{1.6pt}
\renewcommand{\arraystretch}{0.82}
\resizebox{\textwidth}{!}{
\begin{tabular}{lccccccccc@{\hspace{6pt}}ccccccccc}
\toprule
\multirow{3}{*}{\textbf{Model}}
& \multicolumn{9}{c}{\textbf{CH}}
& \multicolumn{9}{c}{\textbf{EN}} \\
\cmidrule(lr){2-10}\cmidrule(lr){11-19}
& \multicolumn{4}{c}{\textsc{HaluReal}}
& \multicolumn{2}{c}{\textsc{HaluFake}}
& \multicolumn{3}{c}{\textsc{HaluAgent}}
& \multicolumn{4}{c}{\textsc{HaluReal}}
& \multicolumn{2}{c}{\textsc{HaluFake}}
& \multicolumn{3}{c}{\textsc{HaluAgent}} \\
\cmidrule(lr){2-5}\cmidrule(lr){6-7}\cmidrule(lr){8-10}
\cmidrule(lr){11-14}\cmidrule(lr){15-16}\cmidrule(lr){17-19}
& \textbf{Acc} & \textbf{UR} & \textbf{SpanF1} & \textbf{CR}
& \textbf{UR} & \textbf{Recall}
& \textbf{Acc} & \textbf{UR} & \textbf{RTAcc}
& \textbf{Acc} & \textbf{UR} & \textbf{SpanF1} & \textbf{CR}
& \textbf{UR} & \textbf{Recall}
& \textbf{Acc} & \textbf{UR} & \textbf{RTAcc} \\
\midrule
Gemini3.1-Pro 
& \textbf{84.35} & 9.40 & \textbf{70.07} & \textbf{82.27} & 44.60 & 75.70 & \textbf{64.55} & \textbf{2.20} & 54.93
& \textbf{84.87} & 16.63 & \textbf{72.95} & \textbf{86.00} & \textbf{93.20} & 53.40 & 72.28 & 1.35 & \textbf{76.16} \\

Gemini3-Flash
& 77.62 & 0.25 & 59.80 & 72.60 & 1.20 & \textbf{96.20} & 50.60 & 0.00 & 54.02
& 81.53 & 0.85 & 60.12 & 72.93 & 12.20 & 93.90 & 72.12 & 0.05 & 72.76 \\

GPT5.4 
& 63.08 & 16.85 & 39.32 & 56.42 & 60.00 & 67.80 & 50.45 & 0.00 & 57.09
& 71.33 & 15.75 & 42.88 & 63.83 & 92.00 & 54.00 & \textbf{72.36} & 0.65 & 73.31 \\

GPT5-Mini 
& 63.35 & \textbf{43.40} & 43.55 & 57.43 & 7.40 & 94.10 & 50.90 & 0.30 & 58.40
& 65.28 & \textbf{55.55} & 42.61 & 57.16 & 6.60 & \textbf{96.70} & 70.73 & 0.85 & 70.24 \\

DeepSeek-V4-Pro 
& 68.57 & 5.45 & 49.33 & 67.96 & 6.80 & 94.00 & 50.00 & 0.00 & 50.10
& 72.10 & 8.50 & 46.99 & 64.57 & 42.20 & 78.90 & 65.80 & 0.00 & 71.83 \\

DeepSeek-V4-Flash 
& 70.90 & 7.50 & 43.13 & 58.18 & 55.00 & 66.50 & 52.15 & 0.00 & 56.92
& 70.97 & 4.85 & 38.01 & 44.20 & 78.40 & 60.60 & 68.45 & 0.00 & 72.91 \\

HY3-Preview
& 62.65 & 19.80 & 39.78 & 58.13 & \textbf{63.60} & 65.00 & 50.00 & 0.00 & 51.00
& 62.20 & 23.10 & 36.88 & 50.25 & 85.00 & 57.50 & 65.76 & 0.30 & 73.16 \\

Llama-3.3-70B-Instruct 
& 59.50 & 9.50 & 19.21 & 45.54 & 28.40 & 64.20 & 45.55 & 0.00 & 57.86
& 63.50 & 15.60 & 25.29 & 40.45 & 58.00 & 70.40 & 70.60 & 0.00 & 70.43 \\

Llama-3.1-8B-Instruct 
& 52.88 & 7.55 & 16.71 & 18.13 & 29.00 & 63.90 & 49.85 & 0.00 & 51.96
& 54.80 & 1.60 & 23.65 & 21.12 & 10.40 & 90.40 & 52.80 & 0.00 & 54.32 \\

Qwen3-235B-A22B 
& 63.12 & 15.85 & 37.99 & 44.53 & 24.60 & 84.10 & 50.50 & 0.00 & 58.01
& 61.75 & 27.10 & 28.33 & 39.81 & 85.40 & 57.30 & 70.25 & \textbf{1.70} & 70.99 \\

Qwen3-30B-A3B 
& 58.03 & 14.65 & 24.94 & 26.61 & 48.80 & 52.80 & 46.05 & 0.00 & \textbf{68.07}
& 57.55 & 43.80 & 20.41 & 15.88 & 91.40 & 52.90 & 66.93 & 0.45 & 69.64 \\

Qwen3-32B 
& 60.22 & 6.45 & 31.03 & 40.40 & 14.20 & 83.10 & 47.50 & 0.00 & 59.66
& 63.68 & 10.65 & 26.63 & 33.23 & 20.20 & 83.10 & 68.80 & 0.00 & 68.84 \\

Qwen3-14B 
& 57.47 & 6.15 & 26.30 & 33.43 & 9.20 & 89.00 & 50.55 & 0.00 & 59.01
& 58.07 & 17.95 & 22.36 & 20.79 & 54.80 & 70.80 & 67.77 & 0.05 & 66.06 \\

Qwen3-8B 
& 52.80 & 0.00 & 23.47 & 22.21 & 0.00 & 91.00 & 54.90 & 0.00 & 61.08
& 53.75 & 0.30 & 25.38 & 14.99 & 2.20 & 95.90 & 66.55 & 0.00 & 65.29 \\
\bottomrule
\end{tabular}
}
\vspace{-0.6em}
\caption{Main results on \textsc{HaluCheck}. All values are percentages. UR is short for UncertainRate, CR is short for CorrectionRate, and RTAcc is short for RootTurnAccuracy. The best result for each metric is highlighted in \textbf{bold}.}
\label{tab:main-results}
\vspace{-1.9em}
\end{table*}

\subsection{Models and Settings}

We evaluate \textsc{HaluCheck} on 14 LLMs across 6 model families. For closed-source models, we include Gemini3.1-Pro, Gemini3-Flash~\citep{google2025gemini3}, GPT5.4, GPT5-Mini~\citep{openai2026gpt54}, DeepSeek-V4-Pro, DeepSeek-V4-Flash~\citep{deepseek2026v4}, and HY3-Preview~\citep{tencent2026hy3}. For open-source models, we include Llama-3.3-70B-Instruct and Llama-3.1-8B-Instruct \citep{grattafiori2024llama}, together with five Qwen3 variants: Qwen3-235B-A22B, Qwen3-30B-A3B, Qwen3-32B, Qwen3-14B, and Qwen3-8B \citep{qwen2025qwen3}. Evaluation prompts are provided in Appendix~\ref{sec:experiment-details}\footnote{To ensure experiment reproducibility, the decoding temperature was set to 0.}. We use the balanced lite test set described in Table~\ref{tab:data-distribution}.

\subsection{Results}

Table~\ref{tab:main-results} presents the main experimental results.
Across both the text and agent tasks, we observe substantial performance gaps among models, highlighting the strong discriminative power of our benchmark. The two tasks emphasize different types of hallucinations: the text task targets factual hallucinations, while the agent task focuses on faithfulness hallucinations, leading to distinct model behaviors.

\paragraph{\textsc{HaluReal}.}

Among all evaluated models, Gemini3.1-Pro achieves the highest detection accuracy of around 84\% and delivers the top performance in localization and correction. Interestingly, it also shows a higher tendency to express uncertainty than Gemini3-Flash.
In contrast, most other models still struggle with hallucination detection, yielding accuracies between 50\% and 70\% on both Chinese and English datasets.
Furthermore, localizing and correcting hallucinations present an even greater challenge. Across almost all models, SpanF1 and CorrectionRate are substantially lower than detection accuracy, highlighting a gap between detecting the presence of a hallucination and accurately localizing and correcting it.

\paragraph{\textsc{HaluFake}.}
Models exhibit distinct uncertainty strategies when facing fabricated entities. In general, stronger models are more likely to express uncertainty about unknown entities, suggesting that they better ``know what they do not know.'' When refusal or uncertainty is allowed, some frontier/pro models adopt a more conservative strategy for fabricated entities. For instance, Gemini3.1-Pro and GPT5.4 show high UncertainRate on English \textsc{HaluFake}, accompanied by relatively lower Recall. In contrast, some lightweight or mini models tend to give direct judgments, resulting in higher Recall and lower UncertainRate, which may also reflect overconfidence in hallucination detection. 
This trade-off suggests that \textsc{HaluFake} evaluates not only whether models can recognize fabricated content, but also how they navigate the delicate balance between helpfulness (attempting to answer) and honesty (admitting ignorance).

\paragraph{\textsc{HaluAgent}.}
\textsc{HaluAgent} is more challenging than text-based tasks: the best sample-level accuracy reaches only 64.55\% in Chinese and 72.36\% in English. 
Unlike factual hallucination detection, which can draw on pre-training knowledge, agent hallucinations require evaluating faithfulness, demanding precise integration of contextual information rather than simple fact-checking. 
In this setting, models rarely express uncertainty, yet their judgments are often unreliable. 
Model performance remains weak on both sample-level detection and root-cause localization, suggesting that identifying whether an agent trajectory contains hallucinations and where it first occurs is difficult. 

Agent trajectories are longer, contain more intermediate states and step dependencies, and require faithfulness judgments over user instructions, reasoning steps, tool calls, and observations. Thus, \textsc{HaluAgent} exposes current models' weaknesses in long-context understanding, process tracking, and error attribution.

Overall, LLMs demonstrate some capability in detecting hallucinations on Twisted Facts, but substantial gaps persist between models, particularly in precise localization and correction. 
Stronger models are better at expressing uncertainty when evidence is insufficient, reflecting greater self-awareness. 
In contrast, agent hallucination poses a far greater challenge: current models struggle with context integration and process tracking, highlighting substantial limitations in detecting and attributing agent hallucinations. Considerable progress is still needed for robust faithfulness in complex, multi-step scenarios.

Additional analyses further validate the practical relevance of \textsc{HaluCheck}. Across 14 evaluated models, benchmark performance shows strong correlations with natural-hallucination detection in factual and agent metrics. These findings suggest that \textsc{HaluCheck} can serve as a useful proxy for evaluating natural-hallucination detection. Furthermore, models that hallucinate less during generation tend to localize hallucinated spans more accurately, suggesting a shared reliance on factual knowledge across generation and detection tasks. See more analysis in Appendices~\ref{sec:natural-validation} and~\ref{sec:generation-detection}.

\section{Discussion}
\label{sec:discussion}

Since hallucination detection depends on the model's internal knowledge and reasoning ability as well as external information, we explore whether explicit thinking, web search integration, and supervised fine-tuning can further improve LLM performance.

\subsection{Impact of Explicit Thinking}

\begin{table*}[t]
\centering
\scriptsize
\setlength{\tabcolsep}{3pt}
\renewcommand{\arraystretch}{1.15}
\begin{tabular}{@{}p{0.115\textwidth}p{0.275\textwidth}p{0.275\textwidth}p{0.275\textwidth}@{}}
\toprule
\textbf{Component} & \textbf{Case 1} & \textbf{Case 2} & \textbf{Case 3} \\
\midrule
\textbf{Input Snippet}
& ``... Samsung TV Plus ... due to its integration with the webOS operating system.''
& ``The 2014 crime drama film starring Tom Hardy and James Gandolfini ... was directed by Micha\"el R. Roskam.''
& ``He is currently Professor ... at McGill University, and Recurrent Visiting professor ... at the Central European University in Prague, Hungary.'' \\
\midrule
\textbf{GroundTruth}
& \textit{Hallu}: Tizen $\rightarrow$ webOS
& \textit{Clean}: no hallucination
& \textit{Hallu}: Queen's University $\rightarrow$ McGill University; Budapest, Hungary $\rightarrow$  Prague, Hungary\\
\midrule
\textbf{Key CoT Content}
& ``\textbf{webOS is actually used by LG}, not Samsung. Samsung uses \textbf{Tizen OS} for their smart TVs.''
& ``\textbf{James Gandolfini died in 2013}, but \textbf{the film was released in 2014}. However, the film was shot before his death and released afterward, \textbf{so he could still appear in it}.''
& ``\textbf{Kymlicka is associated with Queen's University}, not McGill. Also, \textbf{Central European University is in Budapest}, Hungary, not `Prague, Hungary'.'' \\
\bottomrule
\end{tabular}
\vspace{-0.6em}
\caption{Representative cases where enabling thinking corrects hallucination detection decisions.}
\label{tab:hallutext-gold-cases}
\vspace{-1.9em}
\end{table*}

\begin{figure}[t]
    \centering
    \includegraphics[width=1\linewidth]{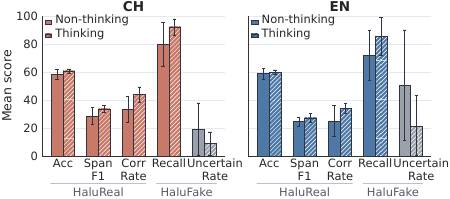}
\vspace{-12pt}
\caption{Effect of thinking mode (Mean $\pm$ std. across five Qwen3 Models).}
    \label{fig:qwen3-thinking}
\vspace{-1.9em}
\end{figure}

Figure~\ref{fig:qwen3-thinking} shows the average performance differences of the Qwen3 series with and without thinking. By enabling thinking, the models improve on \textsc{HaluReal}: they identify hallucinated spans more accurately and better correct erroneous content. On \textsc{HaluFake}, thinking mode increases Recall while reducing UncertainRate, suggesting that reasoning makes models more confident and more willing to provide definitive judgments.

Table~\ref{tab:hallutext-gold-cases} shows representative cases where thinking mode changes the model's hallucination judgments. The examples indicate that thinking enables models to explicitly check entity attributes, temporal relations, institutional affiliations, and geographic consistency before making a final decision. By recalling and verifying relevant knowledge, thinking helps models recover missed hallucinations, avoid false positives, and provide more accurate span-level corrections.

\subsection{Impact of Web Search Integration}

\begin{figure}[t]
    \centering
    \includegraphics[width=1\linewidth]{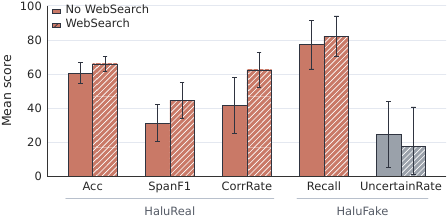}
\vspace{-12pt}
\caption{Effect of web search (Mean $\pm$ std. across five Qwen3 Models).}
    \label{fig:web-search}
\vspace{-1.9em}
\end{figure}

Web search \citep{cheng2024small} and RAG \citep{lewis2020retrieval} are commonly used to mitigate factual hallucinations. 
As shown in Figure~\ref{fig:web-search}, integrating web search results\footnote{We obtain the web search results from the GLM Web Search API: \url{https://docs.bigmodel.cn/cn/guide/tools/web-search}} consistently enhances model performance in hallucination detection, localization, and correction, while also reducing the rate of uncertain responses. 
The improvement is generally greater than that achieved by enabling explicit reasoning alone, highlighting the direct value of external knowledge for factual verification. However, the gains are not uniform: stronger models see marginal improvement, whereas weaker models benefit more yet may still lag behind top models without retrieval. 
This suggests that retrieval augmentation cannot fully substitute for a model's inherent factual understanding and information integration capabilities. In particular, the effectiveness of web search depends on both the quality of retrieved evidence and the model's ability to filter and utilize this information. Retrieved evidence can be incomplete, noisy, or poorly matched, and weaker models may struggle to make use of it effectively. Thus, while web search can help mitigate factual hallucinations, its overall benefit is limited by both retrieval quality and the model's external knowledge integration.

\subsection{Impact of Supervised Fine-tuning}

For supervised fine-tuning (SFT), we sample 5,000 paired positive-negative instances from the full \textsc{HaluCheck} set and verify that there is no overlap with the balanced lite set used as our test set. We fine-tune Qwen3-8B and Qwen3-14B using LoRA~\citep{hu2022lora} on this data. See detailed settings in Appendix~\ref{sec:sft-details}.
As shown in Figure~\ref{fig:sft-generalization}, SFT yields consistent performance gains in hallucination detection, localization, and correction. Given that the fine-tuning and test sets share no knowledge or instance overlap, we attribute these improvements primarily to the model's enhanced ability to capture hallucination patterns and to detect subtle inconsistencies in the data.

To further evaluate the generalization ability of the fine-tuned models, we construct external test sets from HaluEval, C-Faith, and AgentHallu\footnote{These datasets have only sample-level labels, so we use binary classification without \textit{Uncertain}.}. Notably, while the fine-tuned models show substantial improvement on the Chinese text tasks, we observe performance drops on English text and agent-based tasks. We hypothesize that this gap is caused by distributional differences between the synthetic training data and the external benchmarks, particularly with respect to error types and contextual complexity. This finding highlights the challenge of achieving robust out-of-distribution generalization and points to the need for more diverse and representative training data for hallucination detection.

\begin{figure}[t]
    \centering
    \includegraphics[width=1\linewidth]{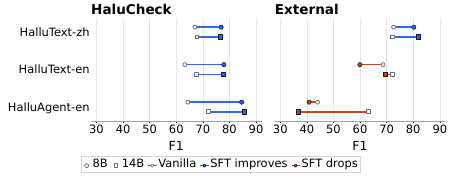}
    \vspace{-10pt}
    \caption{Effect of LoRA Fine-tuning.}
    \label{fig:sft-generalization}
\vspace{-1.4em}
\end{figure}

\section{Conclusions}
This paper investigates whether LLMs can reliably detect hallucinations after generation and proposes \textsc{HaluCheck}, a bilingual, fine-grained benchmark covering factual hallucinations in text and faithfulness hallucinations in agent trajectories. Unlike benchmarks that mainly measure whether models generate hallucinations, \textsc{HaluCheck} evaluates whether models can identify existing hallucinations, localize ``what is wrong,'' and determine ``where the error starts.'' Experiments show that current LLMs have some hallucination detection ability, but remain limited in span localization, correction, and agent root-step attribution. Reasoning mode, web search, and supervised fine-tuning can improve detection performance to some extent, but their gains are constrained by model capability, evidence quality, and training data coverage, indicating that reliable hallucination detection remains an open challenge.

\par\label{sec:main-content-end}
\section*{Limitations}
Although \textsc{HaluCheck} provides a controlled evaluation of hallucination detection, several limitations remain. First, although constructed hallucinations may differ from naturally occurring ones, the observed correlations in evaluation metrics support their use as a practical evaluation proxy. Second, the benchmark focuses on representative factual-text and agent scenarios, leaving broader real-world settings for future exploration. Third, an independent human detection baseline is challenging due to the extensive factual verification and trajectory tracing required, although we have conducted manual verification to ensure data quality. Finally, potential model-based biases are a common challenge in LLM-assisted generation and evaluation, which we address through multi-model validation and cross-judge evaluations. Future work will further expand scenario coverage and strengthen real-world validation.

\section*{Ethical Considerations}

This work focuses on hallucination detection and uses only publicly available open-source datasets, without involving private, sensitive, or personally identifiable information. The benchmark is constructed through controlled synthetic perturbations, with rule-based checks and multi-model validation applied to improve data quality and reduce potential bias. Since the dataset contains hallucinated texts and perturbed agent trajectories, it should be used only for reliability evaluation and safety research, not for spreading misinformation or unsafe agent behavior. Code repository: \halucheckproject.

\section*{Acknowledgments}
This research was supported by the Tencent Basic Platform Technology Rhino-Bird Focused Research Program.

\bibliography{custom}

\appendix
\section{Data Sources and Licenses}
This section summarizes the data sources used to construct \textsc{HaluText}, \textsc{HaluAgent}, and the external test sets, and documents the corresponding license information. We distinguish perturbation sources from evaluation-only resources to clarify data provenance, reproducibility, and compliance with the usage terms of each source dataset.

\subsection{Data Sources}
\paragraph{Perturbation Data Sources.}
For \textsc{HaluText}, we use both English and Chinese Wikipedia, together with factual QA datasets including HotPotQA~\citep{yang2018hotpotqa}, PopQA~\citep{mallen2023llm_memorization}, HalluQA~\citep{cheng2023evaluating}, and CMRC2018~\citep{cui-emnlp2019-cmrc2018}. For \textsc{HaluAgent}, we use agent SFT data from AgentFLAN~\citep{chen2024agent}, AgentBank~\citep{song2024agentbank}, AgentTraj-L~\citep{xi2025agentgym}, ToolACE~\citep{liu2025toolace}, and MSAgent-Bench~\citep{li-etal-2023-modelscope}. All hallucinated samples are constructed by perturbing real, non-hallucinated data, so that each hallucination can be paired with its original factual counterpart.
Chinese and English examples are collected from language-specific source datasets and constructed independently, rather than translated from one language to the other.

\paragraph{External Test Set Sources.}
To evaluate the out-of-distribution generalization ability of the synthetic data constructed in this work, we build external test sets from public datasets, including HaluEval~\citep{li2023halueval}, C-Faith~\citep{zhang2025c}, and AgentHallu~\citep{liu2026agenthallu}. For HaluEval, we use the provided evidence and correct answers, and prompt an LLM to generate corresponding hallucinated answers. The correct answers are used as non-hallucinated samples, while the generated incorrect answers are used as hallucinated samples, with 1,000 samples for each class. For C-Faith, we directly use the \texttt{correct\_answer} field as non-hallucinated samples and the \texttt{hallucinated\_answer} field as hallucinated samples, again sampling 1,000 instances for each class. For \textsc{AgentHallu}, we reuse its original annotations, which contain 693 samples in total, including 443 hallucinated samples and 250 non-hallucinated samples. The internal test set is the same as that used in the main experiments.

\subsection{Data Licenses}
We use publicly available datasets and follow their stated licenses and attribution requirements.

\begin{itemize}
    \item \textbf{\textsc{HaluText}.}
    \textsc{HaluText} uses Wikipedia passages and factual QA benchmarks. Wikipedia passages are based on Wikipedia text, which is licensed under CC BY-SA 4.0. HotPotQA~\citep{yang2018hotpotqa} is released under CC BY-SA 4.0, including its processed Wikipedia corpus. PopQA~\citep{mallen2023llm_memorization} is released through the official Adaptive Retrieval repository under the MIT license, with its underlying Wikidata-derived facts available under CC0. HalluQA~\citep{cheng2023evaluating} is released under Apache-2.0. CMRC2018~\citep{cui-emnlp2019-cmrc2018} is released under CC BY-SA 4.0.

    \item \textbf{\textsc{HaluAgent}.}
    \textsc{HaluAgent} is constructed from real agent trajectories drawn from publicly released agent datasets and benchmarks. AgentFLAN~\citep{chen2024agent} is released under Apache-2.0. AgentBank~\citep{song2024agentbank} is released under Apache-2.0. AgentTraj-L from AgentGym~\citep{xi2025agentgym} is publicly hosted on Hugging Face; its dataset card does not specify a separate data license, while the accompanying AgentGym release is MIT-licensed. ToolACE~\citep{liu2025toolace} is released under Apache-2.0. MSAgent-Bench~\citep{li-etal-2023-modelscope} is released under Apache License 2.0.

    \item \textbf{External test sets.}
    HaluEval~\citep{li2023halueval} is released under the MIT license. C-Faith~\citep{zhang2025c} is publicly released through its official repository, which does not specify a separate license file. AgentHallu~\citep{liu2026agenthallu} is released under CC BY 4.0.
\end{itemize}

\section{Details of HaluCheck Construction}
This section describes the construction process of \textsc{HaluCheck}, including model setup, the multi-model synthesis pipeline, quality control, and human verification. It explains how text, QA, and agent sources are routed, perturbed, filtered, and validated to produce diverse, traceable, and reliable hallucination annotations.

\subsection{Setup}
\paragraph{Model selection.}
For synthetic data generation, we use three frontier open-sourced large models: Qwen3.5-397B-A17B~\citep{qwen2026qwen35}, GLM4.7~\citep{5team2025glm45agenticreasoningcoding}, and Llama4-Maverick~\citep{meta2025llama4}. All three models are deployed locally with FP8 quantization through SGLang on a single node with 8 H20 GPUs. During the synthesis stage, each model independently perturbs a subset of the source data. During the quality-control stage, every generated sample is evaluated by all three models in turn. A sample is retained in the final dataset only if all three models assign quality scores above the predefined threshold.

The use of multiple model families is intended to reduce family-specific artifacts in the generated perturbations. We also vary source texts, perturbation targets, and perturbation strategies so that the final labels are not determined by any single generator family.

\subsection{Construction Pipeline}
\label{sec:multi-model}
\label{sec:human-verification}

To ensure diversity and quality, we build a multi-model collaborative data synthesis pipeline consisting of three stages: data filtering and routing, perturbation, and quality control.

\paragraph{Text and QA Processing.}
We preprocess the different data sources according to their formats and task types. For Wikipedia, we remove entries with abnormal text length or paragraph counts, use the first paragraph as the core factual passage, and treat the remaining paragraphs as supplementary context. For each entry, we randomly select either the core paragraph or a supplementary paragraph as the perturbation source. We then perform NER and sample entities under three perturbation-ratio settings, 0.1, 0.25, and 0.6. These ratio settings control the number of perturbed entities and are randomly combined with the Obvious, Moderate, and Subtle perturbation-degree prompts. For fabricated-entity cases, we sample Wikipedia core paragraphs and prompt models to imitate their encyclopedic style while generating entirely fictional entity descriptions. For QA data, we filter questions with overly short or long answers, and provide the model with the question, correct answer, incorrect answer, and supporting evidence when available.

\paragraph{Agent Trajectory Selection.}
For agent data, we select trajectories aligned with text-based agent reasoning and tool use. AgentFLAN contributes ToolBench-style ReAct, CoT, and instruction-following trajectories; AgentTraj-L contributes weather, movie, todo, and SQLGym trajectories; ToolACE provides multi-turn tool-use examples; and MSAgent-Bench provides Chinese ModelScope plugin-based dialogues. For AgentBank, we treat each subset as an independent source for provenance tracking and flexible filtering, and retain text-oriented subsets covering code generation, mathematical reasoning, search-based QA, bash and SQL execution, web interaction, and online shopping. We exclude embodied or visually grounded subsets such as ALFRED, ALFWorld, IQA, and Rearrange, since their physical actions and visual observations are less aligned with text-based agent hallucinations. To control trajectory quality and context length, we keep ToolACE examples with 2--10 turns, constrain long instances to a maximum budget of 8,192 tokens, and use a fixed random seed of 42 for reproducible sampling.

\paragraph{Perturbation Generation.}
The \textit{perturbation} stage uses three LLMs to independently generate hallucinated samples according to predefined perturbation strategies and perturbation degrees. For \textsc{HaluText}, perturbations include substitution, displacement, and fabrication over factual entities or evidence. For \textsc{HaluAgent}, we perturb only one turn in each trajectory to introduce a planning, tool-use, or perception hallucination, and then update the subsequent trajectory to keep the interaction coherent. Each generated sample preserves metadata such as the original content, perturbed content, perturbation position, hallucination type, and perturbation degree, enabling traceable span-level or turn-level annotation.

\paragraph{Quality Control.}
The \textit{quality control} stage combines rule-based and model-based checks. Rule-based checks verify whether model outputs follow the predefined schema, whether the original and perturbed fields correspond correctly, whether factual and hallucinated samples are properly paired, and whether agent perturbations keep the original trajectory length unchanged. Model-based quality control asks multiple models to score each generated sample in terms of perturbation validity, strategy consistency, fluency, coherence, causal consistency, and naturalness. A sample is retained in the final dataset only if all model-based scores meet the predefined quality threshold. We further validate whether the controlled perturbations in \textsc{HaluCheck} provide a useful proxy for naturally occurring hallucinations. We compare the fluency of hallucinated and clean texts using Qwen3-8B-Base perplexity. As shown in Table~\ref{tab:ppl-fluency}, hallucinated texts have perplexity values close to clean texts in both languages, suggesting that the perturbations do not create visibly unnatural artifacts.

In addition to the automatic quality and fluency checks, we manually inspect 100 sampled instances from each subset to assess label reliability. Table~\ref{tab:human-verification} reports both sample validity and perturbation validity. The results show high sample validity across subsets, while also revealing that agent perturbations are more challenging to construct than text perturbations. This human inspection is designed to verify known labels rather than to establish a from-scratch human baseline. A human baseline for this task would require annotators to independently detect subtle factual errors and trace multi-step agent trajectories without access to the injected labels, which is substantially more demanding than validating label correctness. We therefore focus on human verification of sample and perturbation validity in this version.

\begin{table}[t]
\centering
\small
\setlength{\tabcolsep}{7pt}
\renewcommand{\arraystretch}{1.08}
\begin{tabular}{lcc}
\toprule
\textbf{Language} & \textbf{Hallucinated} & \textbf{Clean} \\
\midrule
CH & 16.74 & 16.67 \\
EN & 12.14 & 11.05 \\
\bottomrule
\end{tabular}
\vspace{-0.3em}
\caption{Qwen3-8B-Base perplexity comparison between hallucinated and clean texts.}
\label{tab:ppl-fluency}
\vspace{-0.8em}
\end{table}

\begin{table}[t]
\centering
\small
\setlength{\tabcolsep}{5pt}
\renewcommand{\arraystretch}{1.08}
\begin{adjustbox}{max width=\columnwidth}
\begin{tabular}{llcc}
\toprule
\textbf{Dataset} & \textbf{Lang.} & \textbf{Sample Validity} & \textbf{Perturbation Validity} \\
\midrule
\textsc{HaluText} & CH & 100.00\% & 97.74\% \\
 & EN & 100.00\% & 98.19\% \\
\textsc{HaluAgent} & CH & 99.00\% & 84.10\% \\
 & EN & 95.00\% & 92.19\% \\
\bottomrule
\end{tabular}
\end{adjustbox}
\vspace{-0.3em}
\caption{Human verification results based on 100 sampled instances from each subset.}
\label{tab:human-verification}
\vspace{-0.8em}
\end{table}

\subsection{Representative Perturbation Cases}
\label{sec:perturbation-cases}

Table~\ref{tab:perturbation_cases} presents representative cases from \textsc{HaluText} and \textsc{HaluAgent}, illustrating the perturbation degrees and the changes made to the original factual content or agent trajectory.

\begin{table*}[t]
\centering
\scriptsize
\renewcommand{\arraystretch}{1.05}

\begin{adjustbox}{max width=\textwidth, max height=0.78\textheight}
\begin{tabular}{C{1.45cm} C{1.15cm} L{6.05cm} L{6.35cm}}
\toprule
\textbf{Benchmark} 
& \textbf{Degree} 
& \multicolumn{1}{c}{\textbf{Hallucinated / Perturbed Case}} 
& \multicolumn{1}{c}{\textbf{Perturbation Description}} \\
\midrule

\multirow{3}{*}{\textsc{HaluReal}}
& Obvious
& Red Axes is a \textbf{Norwegian} electronic and indie music duo formed in \textbf{1985}. The duo is based in \textbf{Reykjavik}.
& Israeli $\rightarrow$ Norwegian; 2010 $\rightarrow$ 1985; Tel Aviv $\rightarrow$ Reykjavik. \\[0.35em]

& Moderate
& Tina Datta is a \textbf{Bangladeshi} actress known for \textbf{Marathi television}. She played Ichcha in \textbf{Naagin} and later joined \textbf{Bigg Boss 13}.
& Indian $\rightarrow$ Bangladeshi; Hindi television $\rightarrow$ Marathi television; \textit{Uttaran} $\rightarrow$ \textit{Naagin}; \textit{Bigg Boss 16} $\rightarrow$ \textit{Bigg Boss 13}. \\[0.35em]

& Subtle
& ``There! I've Said It Again'' is a popular song written and published by Redd Evans and \textbf{Bob Russell} in \textbf{1942}. In \textbf{late 1945}, Vaughn Monroe and his Orchestra released \textbf{Victor 20-1638}... and a total run of \textbf{28 weeks}. It finished 1946 as the no. 5 record of the year, \textbf{a ranking explicitly confirmed in the December 1946 issue of Variety magazine which cited the song's unique modulation as the primary driver of its longevity}.
& David Mann $\rightarrow$ Bob Russell; 1941 $\rightarrow$ 1942; early 1945 $\rightarrow$ late 1945; Victor 20-1637 $\rightarrow$ Victor 20-1638; 29 weeks $\rightarrow$ 28 weeks; unsupported \textit{Variety} citation added. \\[0.35em]

\midrule
\multirow{1}{*}{\textsc{HaluFake}}
& -
& Ceremonies to honor the lunar cycles were formally established in the Kingdom of Virell in 1873, marking a turning point in the region's cultural and spiritual practices. The practice centered around the Order of Selvath, an ancient sect that emerged during the 17th-century Enlightenment Period and quickly became the foremost authority on celestial navigation and ritual astronomy.
& This case is totally fabricated. \\

\midrule
\multirow{3}{*}{\textsc{HaluAgent}}
& Obvious
& \textit{Tool-use hallucination.} Perturbed root: ``The system has now detected a prompt asking for confirmation to proceed with the destination search ... \textbf{Selection: E Action: CLICK}.''
& Original root selects ``Next / Find Your Destination'' with ``Selection: F Action: CLICK'' $\rightarrow$ perturbed root selects ``YES'' with ``Selection: E Action: CLICK.'' \\[0.35em]

& Moderate
& \textit{Perception hallucination.} Perturbed root: ``The result shows the author is Yessenia Barrows \textbf{with an ID of 8}.''
& Observation returns ``[(Yessenia, Barrows, 5)]'' and original root says ID 5 $\rightarrow$ perturbed root reports ID 8. \\[0.35em]

& Subtle
& \textit{Plan hallucination.} Perturbed root: ``Since he is an Australian actor, I should \textbf{prioritize checking his roles in Australian-produced science fiction films first}, as local actors often star in domestic genre productions. Action: search[Rob Jenkins movies].''
& Original root only plans to search for Rob Jenkins' movies and identify the 2014 science fiction thriller $\rightarrow$ perturbed root injects an Australia-produced-film search priority and changes the search strategy. \\

\bottomrule
\end{tabular}
\end{adjustbox}
\vspace{-0.6em}
\caption{Representative perturbation cases from \textsc{HaluText} and \textsc{HaluAgent} across three perturbation degrees. Hallucinated fragments are highlighted in \textbf{bold}. For \textsc{HaluAgent}, perturbation types are not tied to specific perturbation degrees.}
\label{tab:perturbation_cases}
\vspace{-1.1em}
\end{table*}

\section{Metric Definitions}
\label{sec:metrics}

This section provides the detailed definitions of the evaluation metrics used in \textsc{HaluCheck}. We evaluate models at three levels: sample-level hallucination detection, span-level localization and correction for \textsc{HaluText}, and root-step localization for \textsc{HaluAgent}.

\subsection{Detailed Metric Definitions}
\paragraph{Sample-level metrics.}
At the sample level, the gold label is binary, indicating whether the input contains hallucinations. Model predictions, however, may take three values: \textit{Hallucinated}, \textit{Non-hallucinated}, and \textit{Uncertain}. For hallucinated samples, predictions of \textit{Hallucinated}, \textit{Non-hallucinated}, and \textit{Uncertain} are counted as $TP$, $FN$, and $U^{+}$, respectively. For non-hallucinated samples, predictions of \textit{Hallucinated}, \textit{Non-hallucinated}, and \textit{Uncertain} are counted as $FP$, $TN$, and $U^{-}$, respectively. The total number of samples is:
\begin{equation}
N = TP + FN + FP + TN + U^{+} + U^{-}.
\end{equation}

We treat \textit{Uncertain} as a conservative prediction between correct and incorrect answers, and assign it a partial score of 0.5 in accuracy and recall. Accuracy measures the overall correctness of sample-level judgments:
\begin{equation}
\mathrm{Acc} =
\frac{TP + TN + 0.5(U^{+}+U^{-})}{N}.
\end{equation}
Recall measures the ability to identify hallucinated samples:
\begin{equation}
\mathrm{Recall} =
\frac{TP + 0.5U^{+}}{TP + FN + U^{+}}.
\end{equation}
We also report the uncertainty rate, which measures how often the model abstains from making a definitive judgment:
\begin{equation}
\mathrm{UncertainRate} =
\frac{U^{+}+U^{-}}{N}.
\end{equation}
This scoring follows a cost-sensitive view in which abstention is treated as an intermediate outcome: it is less informative than a correct prediction, but preferable to an unsupported incorrect prediction.

\paragraph{Span-level metrics for \textsc{HaluReal}.}

For \textsc{HaluReal}, we evaluate not only whether a model detects hallucinations at the sample level, but also whether it can localize hallucinated spans and correct them. Since the dataset does not exhaustively annotate all non-hallucinated spans, $\mathrm{TN}_{\mathrm{span}}$ is not defined.

In our final evaluation, span matching and correction assessment are performed by an \textbf{LLM judge}. Given the ground-truth hallucinated spans and the model-predicted spans, the judge identifies their correspondence, determines whether each predicted span refers to the same hallucinated content as a gold span, and assesses whether the proposed correction is correct. It then outputs a structured result containing $\mathrm{TP}_{\mathrm{span}}$, $\mathrm{FP}_{\mathrm{span}}$, $\mathrm{FN}_{\mathrm{span}}$, $\#\mathrm{MatchedGT}$, and $\#\mathrm{CorrectSpan}$. A predicted span is counted as $\mathrm{TP}_{\mathrm{span}}$ if it matches a gold hallucinated span, and as $\mathrm{FP}_{\mathrm{span}}$ otherwise. A gold hallucinated span not matched by any prediction is counted as $\mathrm{FN}_{\mathrm{span}}$. $\#\mathrm{MatchedGT}$ denotes the number of gold spans matched by at least one prediction, and $\#\mathrm{CorrectSpan}$ denotes the number of matched hallucinated spans whose corrections are also judged correct. We use Qwen3.5-397B-A22B, locally deployed in FP8, as the LLM judge.

Sample-level errors are propagated to span-level statistics. If a factual sample is predicted as hallucinated, all predicted spans are counted as $\mathrm{FP}_{\mathrm{span}}$. If a hallucinated sample is predicted as non-hallucinated, all gold hallucinated spans are counted as $\mathrm{FN}_{\mathrm{span}}$.

Span-level precision, recall, and F1 are computed as:
\begin{equation}
\mathrm{Precision}_{\mathrm{span}} =
\frac{\mathrm{TP}_{\mathrm{span}}}
{\mathrm{TP}_{\mathrm{span}} + \mathrm{FP}_{\mathrm{span}}},
\end{equation}
\begin{equation}
\mathrm{Recall}_{\mathrm{span}} =
\frac{\#\mathrm{MatchedGT}}{\#\mathrm{GT}},
\end{equation}
\begin{equation}
\mathrm{F1}_{\mathrm{span}} =
\frac{2 \cdot \mathrm{Precision}_{\mathrm{span}} \cdot \mathrm{Recall}_{\mathrm{span}}}
{\mathrm{Precision}_{\mathrm{span}} + \mathrm{Recall}_{\mathrm{span}}}.
\end{equation}

For correction evaluation, we only consider predicted spans that are correctly matched to gold hallucinated spans. The correction accuracy is:
\begin{equation}
\mathrm{Acc}_{\mathrm{corr}} =
\frac{\#\mathrm{CorrectSpan}}{\mathrm{TP}_{\mathrm{span}}}.
\end{equation}

This metric is computed only over correctly localized hallucinated spans, and therefore reflects the model's ability to repair errors after successful localization.

\paragraph{Root-step localization for \textsc{HaluAgent}.}
For \textsc{HaluAgent}, in addition to sample-level hallucination detection, we evaluate whether the model can identify the earliest hallucinated step in the agent trajectory. This metric is computed only for hallucinated samples that the model correctly predicts as hallucinated. Let $\mathrm{RootRight}$ denote the number of such samples where the earliest hallucinated step is correctly localized. The root-step localization accuracy is:
\begin{equation}
\mathrm{RootTurnAcc} =
\frac{\mathrm{\#RootRight}}{TP}.
\end{equation}
This metric measures whether the model can attribute the hallucination to its root cause rather than merely detecting that an error exists somewhere in the trajectory.

\subsection{Robustness of Evaluation Protocols}
\paragraph{Rule-based Span Evaluation.}
We also implement a deterministic \textbf{rule-based span-matching evaluator} as an alternative protocol. This evaluator follows the same accounting of true positives, false positives, false negatives, and corrected spans, but replaces semantic judgment with character-level span matching and normalized string matching. The procedure is shown in Algorithm~\ref{alg:rule-span-eval}.

\begin{algorithm}[t]
\caption{Rule-based Span Evaluation for \textsc{HaluReal}.}
\label{alg:rule-span-eval}
\begin{algorithmic}[1]
\Require Gold label $y$, predicted label $\hat{y}$, gold spans $G$, predicted spans $P$
\Ensure $\mathrm{TP}_{\mathrm{span}}$, $\mathrm{FP}_{\mathrm{span}}$, $\mathrm{FN}_{\mathrm{span}}$, $\#\mathrm{MatchedGT}$, $\#\mathrm{CorrectSpan}$
\State Initialize $\mathrm{TP}_{\mathrm{span}} \gets 0$, $\mathrm{FP}_{\mathrm{span}} \gets 0$, $\#\mathrm{MatchedGT} \gets 0$, $\#\mathrm{CorrectSpan} \gets 0$
\If{$y=\textsc{Clean}$ and $\hat{y}=\textsc{Hallu}$}
    \State $\mathrm{FP}_{\mathrm{span}} \gets |P|$
    \State \Return
\EndIf
\If{$y=\textsc{Hallu}$ and $\hat{y}=\textsc{Clean}$}
    \State $\mathrm{FN}_{\mathrm{span}} \gets |G|$
    \State \Return
\EndIf
\ForAll{predicted span $p \in P$}
    \If{$p$ matches any gold span $g \in G$}
        \State $\mathrm{TP}_{\mathrm{span}} \gets \mathrm{TP}_{\mathrm{span}} + 1$
        \If{the correction of $p$ matches the gold factual content}
            \State $\#\mathrm{CorrectSpan} \gets \#\mathrm{CorrectSpan} + 1$
        \EndIf
    \Else
        \State $\mathrm{FP}_{\mathrm{span}} \gets \mathrm{FP}_{\mathrm{span}} + 1$
    \EndIf
\EndFor
\ForAll{gold span $g \in G$}
    \If{$g$ matches any predicted span $p \in P$}
        \State $\#\mathrm{MatchedGT} \gets \#\mathrm{MatchedGT} + 1$
    \EndIf
\EndFor
\State $\mathrm{FN}_{\mathrm{span}} \gets |G| - \#\mathrm{MatchedGT}$
\end{algorithmic}
\end{algorithm}

In the rule-based evaluator, a predicted span and a gold span are considered matched if their character intervals overlap or one contains the other. Correction correctness is determined by normalized string matching between the predicted correction and the gold factual content.

\paragraph{Rule-based and LLM-judge Correlation.}
To examine whether the choice of span-matching protocol affects the overall trend, we compare the rule-based evaluator with the LLM judge. As shown in Table~\ref{tab:rule-llm-correlation}, the two protocols are positively correlated across all metrics. The LLM judge gives lower absolute scores because it applies stricter semantic matching and correction assessment, but the strong positive correlations indicate that the main trends are robust.

\begin{table}[t]
\centering
\small
\setlength{\tabcolsep}{5pt}
\renewcommand{\arraystretch}{1.08}
\begin{tabular}{@{}lcccc@{}}
\toprule
\textbf{Metric} & \textbf{Rule} & \textbf{LLM} & \textbf{$r$} & \textbf{$\rho$} \\
\midrule
Span-P   & 42.9 & 34.9 & 0.965 & 0.915 \\
Span-R   & 48.6 & 41.5 & 0.975 & 0.964 \\
Span-F1  & 44.4 & 36.8 & 0.967 & 0.948 \\
CorrRate & 62.6 & 47.1 & 0.645 & 0.635 \\
\bottomrule
\end{tabular}
\vspace{-0.3em}
\caption{Correlation between rule-based span matching and LLM-judge evaluation on \textsc{HaluReal}.}
\label{tab:rule-llm-correlation}
\vspace{-1.1em}
\end{table}

To further examine whether the LLM judge introduces model-specific bias, we repeat span-level evaluation with GLM4.7 and compare it with the Qwen3.5 judge used in the main results. Table~\ref{tab:judge-robustness} shows that model-level metrics remain highly correlated across the two judges, indicating that relative trends are robust even though absolute scores differ slightly.

\begin{table}[t]
\centering
\small
\setlength{\tabcolsep}{4pt}
\renewcommand{\arraystretch}{1.08}
\begin{tabular}{@{}lcccc@{}}
\toprule
\textbf{Metric} & \textbf{$r$} & \textbf{$\rho$} & \textbf{Qwen3.5} & \textbf{GLM4.7} \\
\midrule
Precision & 0.9948 & 0.9895 & 0.3520 & 0.3089 \\
Recall & 0.9704 & 0.9862 & 0.4238 & 0.3523 \\
F1 & 0.9892 & 0.9886 & 0.3737 & 0.3208 \\
Correction Acc. & 0.9787 & 0.9780 & 0.4740 & 0.4311 \\
\bottomrule
\end{tabular}
\vspace{-0.3em}
\caption{Robustness of span-level evaluation across Qwen3.5 and GLM4.7 judges.}
\label{tab:judge-robustness}
\vspace{-1.1em}
\end{table}

\section{Details of the Main results}
This section provides additional details for the main experiments, including the evaluation prompts, model-access settings, and supplementary performance analyses. It clarifies how \textsc{HaluText} and \textsc{HaluAgent} are evaluated and further breaks down model behavior across perturbation degrees and agent hallucination categories.

\subsection{Experiment Settings}
\label{sec:experiment-details}

For \textsc{HaluText}, we use task-specific prompts to evaluate model outputs on different subsets of \textsc{HaluCheck}. Below we provide the prompt used for hallucination detection and correction in \textsc{HaluText}, which is shared by both \textsc{HaluReal} and \textsc{HaluFake}. For \textsc{HaluFake}, we only use the sample-level hallucination label during evaluation and ignore the correction fields. The prompt instructs the model to determine whether the input text contains hallucinations, localize erroneous spans verbatim, and provide factual corrections when possible. The model is required to return a JSON object with a hallucination label and a list of detected hallucinated spans and corrections.

For \textsc{HaluAgent}, we use a separate prompt tailored to multi-turn agent trajectories. Unlike \textsc{HaluText}, the input includes both the conversation history and the tool definitions available to the agent. The prompt asks the model to detect hallucinations that conflict with user intent, tool definitions, or the trajectory context, and to identify the earliest hallucinated turn as the root hallucination. If later turns inherit errors from this root, they are marked as propagation. The model is required to return a JSON object containing the hallucination label and a list of corrections, where exactly one correction entry should be marked with \texttt{is\_root: true} when hallucinations are detected.

For closed-source models, we access them through their corresponding APIs. Open-source models are deployed locally. For closed-source models that support reasoning mode, we set \texttt{reasoning\_effort} to \texttt{medium}.

\subsection{Model Performance by Perturbation Degree}
\label{sec:appendix-difficulty}
\begin{figure}[t]
    \centering
    \includegraphics[width=\columnwidth]{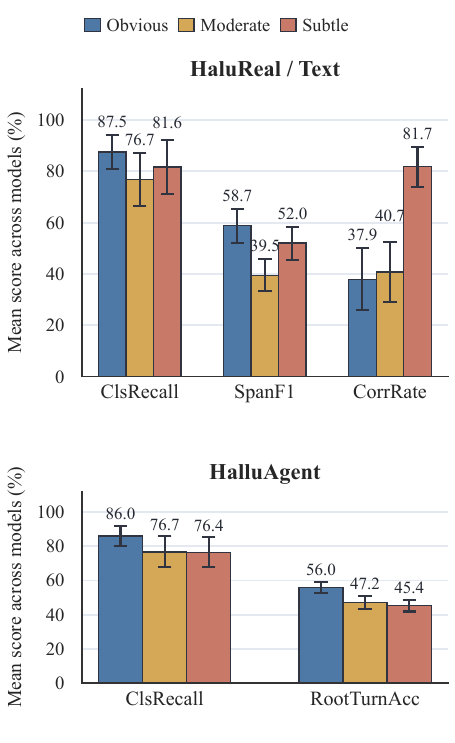}
    \vspace{-17.5pt} 
    \vspace{-12pt}
    \caption{Average model performance across perturbation degrees on HaluReal/Text and HaluAgent. Error bars denote one standard deviation across models.}
    \label{fig:perturbation-degree-performance}
\vspace{-1.1em}
\end{figure}

\begin{figure}[t]
    \centering
    \includegraphics[width=\columnwidth]{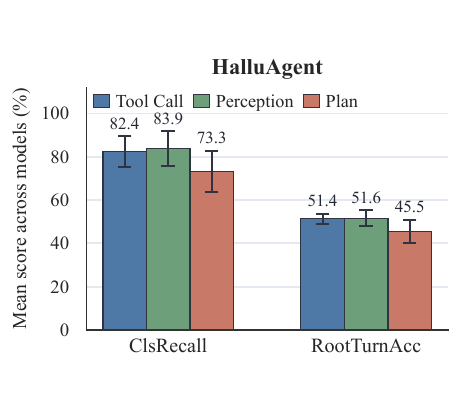}
    \vspace{-21.5pt} 
    \vspace{-12pt}
    \caption{Average HaluAgent performance across perturbation categories. Error bars denote one standard deviation across models.}
    \label{fig:agent-type-performance}
\vspace{-1.1em}
\end{figure}

Figure~\ref{fig:perturbation-degree-performance} shows that models achieve relatively high sample-level hallucination recall, but their fine-grained localization or attribution ability remains substantially weaker. On HaluReal/Text, ClsRecall remains high across perturbation degrees, reaching 87.5\% on Obvious cases, 76.7\% on Moderate cases, and 81.6\% on Subtle cases. However, SpanF1 is much lower, especially on Moderate cases, where it drops to 39.5\%. CorrRate also varies across perturbation degrees, with Subtle cases showing the highest correction rate at 81.7\%, suggesting that classification performance and correction performance do not always align. On HaluAgent, the gap between detection and root-turn attribution is clearer: ClsRecall remains around 76--86\%, whereas RootTurnAcc drops from 56.0\% on Obvious cases to 47.2\% and 45.4\% on Moderate and Subtle cases, respectively.

Figure~\ref{fig:agent-type-performance} further reveals that HaluAgent errors are not equally difficult across perturbation categories. Perception and tool-call perturbations obtain similar ClsRecall scores, at 83.9\% and 82.4\%, respectively, while plan perturbations are harder to detect, with ClsRecall decreasing to 73.3\%. The same trend appears in RootTurnAcc: tool-call and perception errors reach 51.4\% and 51.6\%, whereas plan errors fall to 45.5\%. These results indicate that current models can often recognize that an agent trajectory contains a hallucination, but they struggle to identify the precise root turn, particularly when the hallucination arises from flawed planning rather than an explicit tool-use or perception error.

\section{Additional Analysis Results}
This section reports supplementary analyses beyond the main benchmark results, including fine-tuning settings, validation against external natural-hallucination sets, and the relationship between factual generation quality and span-level hallucination localization. These analyses provide additional evidence about the robustness, transferability, and diagnostic value of \textsc{HaluCheck}.

\subsection{Fine-tuning Details}
\label{sec:sft-details}

\paragraph{Experiment Settings.}
We fine-tune Qwen3-8B and Qwen3-14B on \textsc{HaluText}-CH, \textsc{HaluText}-EN, and \textsc{HaluAgent}-EN using LoRA-SFT with the ms-swift framework. All experiments are conducted on a single node with 8 H20 GPUs, using DeepSpeed ZeRO-2 and bf16 precision. For each of the three training sets, we sample 2,500 positive and 2,500 negative examples from the full synthetic dataset, subject to text-length constraints, and ensure strict de-duplication from the Lite internal evaluation set. The fine-tuning prompts are derived from the baseline detection prompts by removing few-shot demonstrations and the uncertain option, so that the model is trained to make direct binary judgments. For Qwen3 fine-tuning, we include the \texttt{<think>\textbackslash n\textbackslash n</think>} markers in the training responses to preserve the model's reasoning format and avoid degrading its reasoning ability.

\begin{table}[t]
\centering
\scriptsize
\setlength{\tabcolsep}{3pt}
\renewcommand{\arraystretch}{0.92}
\begin{tabular}{p{0.30\linewidth}p{0.62\linewidth}}
\toprule
\textbf{Item} & \textbf{Value} \\
\midrule
Framework / optimizer & ms-swift (\texttt{swift sft}) + DeepSpeed ZeRO-2 / AdamW \\
Tuner & LoRA: rank 16, alpha 32, dropout 0.05 \\
LoRA modules & all-linear; bias none \\
Precision & bf16; gradient checkpointing \\
Per-device batch & train 2 / eval 2 \\
Gradient accumulation & 4 \\
Global batch size & 64 \\
Epochs & 3 \\
LR scheduler & cosine; warmup ratio 0.05 \\
Weight decay / max grad norm & 0.0 / 1.0 \\
AdamW $(\beta_1,\beta_2,\epsilon)$ & $(0.9, 0.95, 1\mathrm{e}{-8})$ \\
Save / eval strategy & per epoch; save total limit 3 \\
Seed & 42 \\
\bottomrule
\end{tabular}
\vspace{-0.4em}
\caption{Hyperparameters for LoRA-SFT experiments.}
\label{tab:sft-hyperparameters}
\vspace{-0.7em}
\end{table}

\subsection{FactScore Reproduction and Data Sources}

To identify major perturbation strategies for factual texts, we first analyze how hallucinations are distributed across entity types. For English, we use the biography subset of the open-source \textsc{HaloGen}~\citep{ravichander2025halogen} hallucination data as the probing source, where Wikipedia-grounded annotations are already provided. For Chinese, we randomly sample entities from Chinese Wikipedia. To ensure coverage across entity popularity, we compute each entity's average monthly page views over the past three years, divide entities into long-tail, medium-popularity, and popular groups using thresholds of 1,000 and 100,000 views, and sample 90 entities evenly from the three groups. We then prompt a set of LLMs to generate entity descriptions. For Chinese generations, we use Qwen3-397B-A37B with FP8 quantization as the verifier and Chinese Wikipedia as the reference database to compute factuality scores. For both English and Chinese data, we apply spaCy to perform named entity recognition over atomic facts, and compute the proportion of entities appearing in hallucinated atomic facts among all extracted entities as the entity-level hallucination rate.

\subsection{Correlation with Natural Hallucination Sets}
\label{sec:natural-validation}

We evaluate all 14 models on natural-hallucination validation sets aligned with \textsc{HaluCheck}. The text validation set contains 400 \textsc{HaluText} instances drawn from UHGEval~\citep{liang2024uhgeval}, FAVA~\citep{mishra2024fava} and HALoGEN~\citep{ravichander2025halogen}. The agent validation set contains 693 English trajectories from AgentHallu~\citep{liu2026agenthallu}. Table~\ref{tab:natural-correlation} reports model-level correlations between \textsc{HaluCheck} and these natural-hallucination sets.

\begin{table*}[t]
\centering
\small
\setlength{\tabcolsep}{5pt}
\renewcommand{\arraystretch}{1.08}
\begin{tabular}{llcc}
\toprule
\textbf{Dataset Pair} & \textbf{Metric} & \textbf{Pearson} & \textbf{Spearman} \\
\midrule
\textsc{HaluText}-CH & Accuracy & $0.8722^{***}$ & $0.8273^{***}$ \\
 & Precision & $0.8376^{***}$ & $0.8889^{***}$ \\
 & Recall & $0.9069^{***}$ & $0.8304^{***}$ \\
 & F1 & $0.9177^{***}$ & $0.9077^{***}$ \\
 & Refusal Rate & $0.9536^{***}$ & $0.9063^{***}$ \\
\midrule
\textsc{HaluText}-EN & Accuracy & $0.9499^{***}$ & $0.9253^{***}$ \\
 & Precision & $0.8465^{***}$ & $0.8823^{***}$ \\
 & Recall & $0.7019^{**}$ & $0.5947^{*}$ \\
 & F1 & $0.8927^{***}$ & $0.8681^{***}$ \\
 & Refusal Rate & $0.9696^{***}$ & $0.9405^{***}$ \\
\midrule
\textsc{HaluAgent}-EN & Accuracy & $0.1652$ & $0.4444$ \\
 & Precision & $0.7224^{**}$ & $0.7582^{**}$ \\
 & Recall & $0.8739^{***}$ & $0.8691^{***}$ \\
 & F1 & $0.6932^{**}$ & $0.6879^{**}$ \\
 & Refusal Rate & $0.7132^{**}$ & $0.6816^{**}$ \\
\bottomrule
\end{tabular}
\vspace{-0.3em}
\caption{Model-level correlations between \textsc{HaluCheck} and natural-hallucination test sets. $^{*}$, $^{**}$, and $^{***}$ denote significance at $p<0.05$, $p<0.01$, and $p<0.001$, respectively.}
\label{tab:natural-correlation}
\vspace{-0.8em}
\end{table*}

\subsection{Relationship between Hallucination Generation and Detection}
\label{sec:generation-detection}

\begin{figure}[t]
    \centering
    \includegraphics[width=1\linewidth]{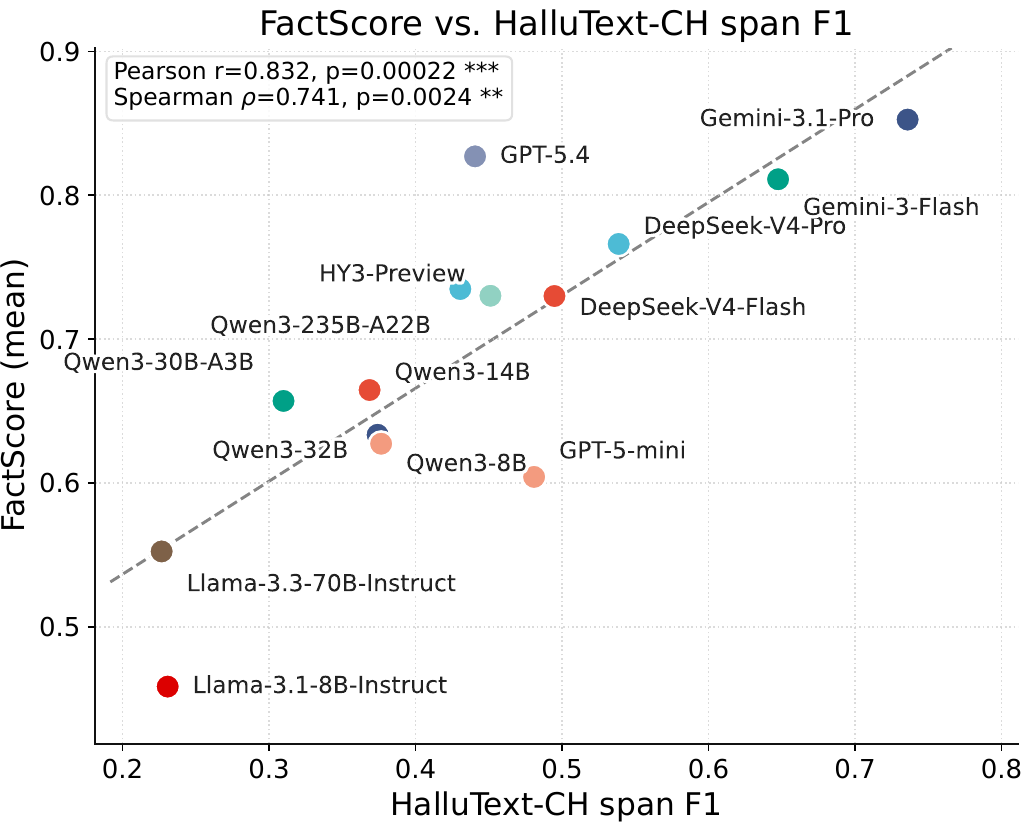}
    \vspace{-12pt}
    \caption{FactScore vs SpanF1}
    \label{fig:factscore-spanf1}
\vspace{-1.9em}
\end{figure}

We evaluate all 14 models by comparing their hallucination generation rates with their detection proficiency. Since the models evaluated in \textsc{HaloGen} are different from those in our experiments, we do not directly compare the two model sets. Figure~\ref{fig:factscore-spanf1} shows the correlation between models' average FactScore on Chinese entity descriptions and their SpanF1 on \textsc{HaluText}-CH. The Pearson correlation is 0.832 and the Spearman correlation is 0.741, both statistically significant at $p<0.01$.

The trend suggests a positive relationship between factual generation ability and hallucination detection ability: models that hallucinate less in their own generations tend to localize hallucinated spans more accurately. This indicates that hallucination detection is partly supported by the model's internal factual knowledge and its ability to verify entity-level information. These results provide additional evidence that hallucination generation behavior and post-generation detection ability are related, while still leaving room for model-specific differences in calibration, evidence use, and span localization.

We hypothesize that this correlation mainly reflects the shared reliance of both tasks on models' memory of factual knowledge. However, factual generation is open-ended, whereas hallucination detection evaluates a fixed input with predefined ground-truth labels and annotations. These different task demands may explain why performance does not fully align across the two tasks.

\section{Summary of Prompt Design}
This section summarizes the prompt suite used for hallucination generation, detection, fine-tuning, and quality control. It documents how the prompts operationalize controlled perturbations, structured outputs, perturbation-degree labels, and automated validation across factual text, QA, fake-content, and agentic trajectory settings.

\paragraph{Prompt Suite Overview.}
The appendix presents a structured prompt suite for hallucination data generation, hallucination detection, fine-tuning, and quality control. The prompts cover multiple scenarios, including Wikipedia-style factual text, question answering, fake-content detection, and agent-based task trajectories.

\paragraph{Synthetic Data Construction.}
For synthetic data construction, the prompts are designed to generate paired samples consisting of a factual reference output and a hallucinated counterpart. In the Wiki passage setting, the main strategy is entity-level perturbation. Given a Wikipedia title, a source passage, and a named-entity list, the prompt validates the entities, determines the number of entities to modify, applies controlled replacements, and records all modifications in a structured format. The generated output includes \texttt{true\_text}, \texttt{hallucinated\_text}, and explicit modification annotations.

\paragraph{Perturbation Degree Design.}
The prompt framework adopts a three-level perturbation-degree design: Obvious, Moderate, and Subtle. Obvious cases introduce easy-to-identify hallucinations through unrelated entity substitutions or large numerical and temporal deviations. Moderate cases use same-category or semantically similar substitutions that require external knowledge to verify. Subtle cases introduce fine-grained and highly believable perturbations, such as closely related entity swaps, fine-grained numerical changes, minor date shifts, or fabricated contextual details.

\paragraph{Fabrication Prompts.}
In addition to entity replacement, the Wiki passage pipeline includes a fabrication-oriented prompt for subtle examples. This prompt adds moderate but false details to an already modified passage, making the hallucination more implicit while preserving fluency and contextual coherence.

\paragraph{QA Prompts.}
For QA data, the prompts generate hallucinated answers from question-answering inputs. Given a question, a correct answer, and optional evidence or incorrect answers, the model is instructed to produce both a correct response and a hallucinated response, together with annotations of the modified factual information.

\paragraph{Agent Prompts.}
For Agent data, the prompts focus on hallucinations in agentic workflows. They cover perception hallucination, plan hallucination, and tool-call hallucination. Perception hallucination changes the assistant's interpretation of tool results while keeping tool outputs unchanged. Plan hallucination modifies reasoning steps or task decomposition. Tool-call hallucination perturbs tool selection or tool parameters. These prompts preserve the original interaction structure while introducing controlled factual or procedural errors.

\paragraph{Quality-control Prompts.}
The quality-control prompts evaluate generated samples along dimensions such as perturbation validity, causal consistency, surface coherence, turn integrity, entity accuracy, text consistency, and fabrication plausibility. These prompts help filter low-quality samples and ensure that hallucinated examples are factually incorrect, linguistically natural, and consistent with the intended perturbation degree.

\paragraph{Design Principles.}
Overall, the prompt suite is built around five core design principles: controlled perturbation, explicit perturbation-degree grading, structured output normalization, self-annotation of modified content, and automated quality assessment. It supports both hallucination generation and hallucination detection across factual text, question answering, fake-content identification, and agentic reasoning scenarios. This design enables the construction of hallucination datasets that are diverse, verifiable, perturbation-aware, and suitable for both evaluation and supervised fine-tuning.

\clearpage
\onecolumn

\begin{figure}[p]
\centering
\begin{promptbox}{HaluText Prompt}
\begin{PromptVerbatim}
# Role
You are an objective [Fact Checker], and you need to verify whether there are hallucinations in the given text.

# Task Description
You will read a text. This text may be completely correct, or it may contain hallucinations (i.e., factual errors, fabricated details). Your tasks are:
1. Detection: Compare the text with real-world historical, scientific, or logical facts to determine if there are hallucinations.
2. Correction: If hallucinations exist, provide factually accurate corrections for each detected hallucination.

# Task Requirements
- If the text content is completely correct, the `corrections` list should be empty.
- If you are unsure whether there are hallucinations, output 'uncertain' in the `has_hallucination` field.
- If the text content contains hallucinations, please complete the task according to the following requirements:
1. Precise localization: The `hallucination` field must extract verbatim the erroneous segment from the original text, pinpointing the specific entity as precisely as possible, without including surrounding punctuation, extraneous words, or using the entire sentence as the detected hallucination.
2. Fact Priority: The corrected content must align with real historical, scientific, or logical facts. If there is information that you consider fabricated or non-existent, delete the part where you believe the hallucination appears, i.e., leave the corresponding "correction" field empty.

# Output Format
Please output ONLY a standard JSON object. Do not include markdown tags or additional explanations. The format is as follows:
{
    "has_hallucination": "true indicates hallucinations exist in the sentence; false indicates no hallucinations exist in the sentence; uncertain indicates you are uncertain whether the sentence has hallucinations",
    "corrections": [
        {
            "hallucination": "The erroneous segment in the original text",
            "correction": "The corrected content; if you consider it fabricated or non-existent information, this field should be empty"
        }
    ]
}

# Examples
Three examples are given below. Please follow the exact output format shown.

## Example 1 (no hallucination)
Input text:
"Paris is the capital of France and is located on the river Seine in the north-central part of the country."
Output:
{"has_hallucination": "false", "corrections": []}

## Example 2 (hallucination present, correctable)
Input text:
"Albert Einstein was born in 1879 in Germany and won the Nobel Prize in Physics in 1925 for his discovery of general relativity."
Output:
{
    "has_hallucination": "true",
    "corrections": [
        {"hallucination": "1925", "correction": "1921"},
        {"hallucination": "general relativity", "correction": "the photoelectric effect"}
    ]
}

## Example 3 (uncertain)
Input text:
"In 1987, the regional novelist Marcus Halloway received the Greater Yorkshire Literary Prize for his debut novel 'Salt and Slate'."
Output:
{"has_hallucination": "uncertain", "corrections": []}

Below is web search information you may refer to (available when web_search is enabled):

Input text:
\end{PromptVerbatim}
\end{promptbox}

\vspace{10pt} 
\vspace{-12pt}
\caption{Prompt used for HaluText}
\label{fig:fact-checking-prompt}
\vspace{-0.8em}
\end{figure}

\clearpage

\begin{figure}[p]
\centering
\begin{promptbox}{HaluAgent Prompt}
\begin{PromptVerbatim}
# Role
You are an objective [AI Agent Conversation Auditor], and you need to verify whether the given multi-turn conversation contains any hallucination.

# Task Description
You will read a multi-turn conversation containing user / assistant / tool roles, along with the list of tool definitions available for this conversation. The conversation may be completely correct, or exhibits other forms of hallucination. Your tasks are:
1. Detection: Check whether the conversation contains any content that conflicts with user intent, conflicts with the tool definitions, or is inconsistent with the context, or other hallucination cases.
2. Root localization: Find the earliest turn where a hallucination occurs. Subsequent chain hallucinations caused by this root should be marked as propagation.
3. Correction: Provide a correction suggestion for each hallucination.

# Task Requirements
- If no hallucination exists in the conversation, the `corrections` list should be empty.
- If you are uncertain whether it is a hallucination, `has_hallucination` should output "uncertain", and the `corrections` list should be empty.
- If hallucinations exist:
    1. Precise localization: The `hallucination` field must extract verbatim the erroneous segment from the original text, without including surrounding extraneous content, and must not use an entire turn as a single hallucination.
    2. Unique root: The `corrections` list must contain exactly ONE entry with `is_root: true`; all others must be `is_root: false`.
    3. Tool definition first: When judging tool-related content, base your judgement strictly on the provided tool definitions, not on your own knowledge.
    4. User intent first: When judging whether the content satisfies the user's need, base your judgement on the original user message.

# Output Format
Please output ONLY a standard JSON object. Do not include markdown tags or additional explanations. The format is as follows:
{"has_hallucination": "true means hallucinations exist in the conversation; false means none exist; uncertain means uncertain","corrections": [{"turn_index": "Turn index where the hallucination occurs, copy from the given input, do not count yourself, output should be a number","is_root": "Boolean, true means root, false means propagation","hallucination": "The erroneous segment from the original text (verbatim)","correction": "Corrected content; if the exact correct value cannot be determined, briefly describe how to correct it","reason": "Brief explanation of why it's an hallucination; if propagation, indicate which root caused it"}]}

# Examples
Three examples are given below. Please follow the exact output format shown.
## Example 1 (no hallucination)
Tools and Conversation:[{"name": "get_weather","description": "Get current weather of a given city","parameters": {"city": {"type": "string", "description": "City name"}}}][{"turn_index": 0, "role": "user", "content": "What's the current weather in Tokyo?"},{"turn_index": 1, "role": "assistant", "tool_calls": [{"name": "get_weather", "arguments": {"city": "Tokyo"}}]},{"turn_index": 2, "role": "tool", "name": "get_weather", "content": {\"city\": \"Tokyo\", \"temp\": 22, \"condition\": \"cloudy\"}"},{"turn_index": 3, "role": "assistant", "content": "It's currently 22°C and cloudy in Tokyo."}]
Output:
{"has_hallucination": "false", "corrections": []}

## Example 2 (hallucination present, root + propagation)
Tools and Conversation:[{"name": "search_flight","description": "Search for flights","parameters": {"origin": {"type": "string", "description": "Origin city"},"destination": {"type": "string", "description": "Destination city"},"date": {"type": "string", "description": "Departure date YYYY-MM-DD"}}}][{"turn_index": 0, "role": "user", "content": "Find me a flight from London to Paris on May 20."},{"turn_index": 1, "role": "assistant", "tool_calls": [{"name": "search_flight", "arguments": {"origin": "London", "destination": "Berlin", "date": "2025-05-20"}}]},{"turn_index": 2, "role": "tool", "name": "search_flight", "content": "[{\"flight\": \"BA8472\", \"price\": 280}]"},{"turn_index": 3, "role": "assistant", "content": "I found flight BA8472 from London to Paris on May 20, priced at \$280."}]
Output:
{"has_hallucination": "true","corrections": [{"turn_index": 1,"is_root": true,"hallucination": "\"destination\": \"Berlin\"","correction": "\"destination\": \"Paris\"","reason": "The user explicitly asked for a flight from London to Paris, but the assistant filled the destination as Berlin in the tool call, violating user intent."},{"turn_index": 3,"is_root": false,"hallucination": "from London to Paris","correction": "Should clarify that the tool actually returned a London-Berlin flight, and the assistant needs to re-invoke the tool with destination=Paris.","reason": "Propagated from the root hallucination at turn 1: the tool actually queried Berlin, but the assistant wrapped the result as a Paris flight."}]}

## Example 3 (uncertain)
Tools and Conversation:[{"name": "lookup_paper","description": "Look up an academic paper","parameters": {"title": {"type": "string"}}}[{"turn_index": 0, "role": "user", "content": "Look up the paper 'Latent Calibration in Sparse Mixture-of-Experts'."},{"turn_index": 1, "role": "assistant", "tool_calls": [{"name": "lookup_paper", "arguments": {"title": "Latent Calibration in Sparse Mixture-of-Experts"}}]},{"turn_index": 2, "role": "tool", "name": "lookup_paper", "content": "{\"found\": false}"},{"turn_index": 3, "role": "assistant", "content": "This paper was published at NeurIPS 2024 by Prof. X's team and proposes a novel calibration method for expert routing."}]
Output:
{"has_hallucination": "uncertain", "corrections": []}

Tools and Conversation:

\end{PromptVerbatim}
\end{promptbox}

\vspace{10pt} 
\vspace{-12pt}
\caption{Prompt used for HaluAgent}
\label{fig:agent-checking-prompt}
\vspace{-0.8em}
\end{figure}

\end{document}